%% file: main.tex
\documentclass{article}

\PassOptionsToPackage{table}{xcolor}
\usepackage{conference,times}
\iclrfinalcopy

\input{math_commands.tex}

\usepackage[utf8]{inputenc}
\usepackage[T1]{fontenc}
\usepackage{graphicx}
\usepackage{comment}
\usepackage{listings}
\usepackage{url}
\usepackage{booktabs}
\usepackage{nicefrac}
\usepackage{microtype}
\usepackage{wrapfig}
\usepackage{algorithm}
\usepackage{algorithmic}
\usepackage{multirow}
\usepackage{subcaption}

\usepackage{xcolor}
\usepackage{array}
\usepackage[most]{tcolorbox}
\usepackage{enumitem}

\usepackage{hyperref}
\definecolor{linkblue}{RGB}{0,70,140}

\hypersetup{
  colorlinks=true,
  linkcolor=linkblue, 
  citecolor=linkblue, 
  urlcolor=linkblue   
}

\definecolor{hdr}{HTML}{16324F}
\providecommand{\best}[1]{\textbf{#1}}
\providecommand{\runr}[1]{\underline{#1}}
\providecommand{\gp}{\rule{0pt}{2.6ex}}
\providecommand{\dset}[1]{\textcolor{hdr}{\bfseries #1}}
\newcommand{\sparsity}[1]{\textcolor{black!55}{\small #1}}

\definecolor{EncoderBlue}{RGB}{58,66,80}
\definecolor{XDMHead}{RGB}{198,217,238}
\definecolor{XDMMetric}{RGB}{223,233,246}
\definecolor{XDMBody}{RGB}{229,237,248}
\definecolor{DDPMHead}{RGB}{234,221,239}
\definecolor{DDPMMetric}{RGB}{245,238,248}
\definecolor{DDPMBody}{RGB}{250,246,252}

\definecolor{StubTeal}{RGB}{27,71,76}
\definecolor{StubMeth}{RGB}{238,240,238}

\definecolor{GrpAHead}{RGB}{197,222,206}
\definecolor{GrpAMet} {RGB}{220,236,226}
\definecolor{GrpABody}{RGB}{233,243,237}

\definecolor{GrpBHead}{RGB}{240,223,199}
\definecolor{GrpBMet} {RGB}{247,237,221}
\definecolor{GrpBBody}{RGB}{251,244,234}

\definecolor{AvgHead} {RGB}{225,190,199}
\definecolor{AvgMet}  {RGB}{242,222,228}
\definecolor{AvgBody} {RGB}{249,237,241}

\definecolor{SpHead}  {RGB}{224,222,217}
\definecolor{SpBody}  {RGB}{243,242,239}

\definecolor{BaseSlate}{RGB}{104,116,118}

\newlength{\encw}
\newlength{\encblk}

\newcommand{\na}{\textcolor{black!45}{--}}
\newcommand{\enc}[1]{%
  \hspace*{-\tabcolsep}%
  {\setlength{\fboxsep}{0pt}%
   \colorbox{StubTeal}{%
     \parbox[c][\encblk][c]{\encw}{\centering
       \textcolor{white}{\textbf{#1}}}}}%
  \hspace*{-\tabcolsep}}

\title{Generative Uncertainty as a Self-supervised Signal for Semantic Similarity Learning}

\author{%
  \makebox[\dimexpr\textwidth-2\tabcolsep\relax][c]{%
    \small
    \begin{tabular}{@{}ccccc@{}}
      Enrico Pallotta\textsuperscript{1,2,*}
      & Sina Raoufi\textsuperscript{1,*}
      & Lars Doorenbos\textsuperscript{1,2}
      & Gianni Franchi\textsuperscript{3}
      & Juergen Gall\textsuperscript{1,2}
    \end{tabular}}\\[1pt]
  \parbox{\dimexpr\textwidth-2\tabcolsep\relax}{\centering
    \normalfont\footnotesize
    \textsuperscript{1}University of Bonn
    \qquad
    \textsuperscript{2}Lamarr Institute for ML \& AI
    \qquad
    \textsuperscript{3}ENSTA Paris, Institut Polytechnique de Paris}\\[1pt]
  \parbox{\dimexpr\textwidth-2\tabcolsep\relax}{\centering
    \normalfont\footnotesize
    \textsuperscript{*}Equal contribution}}
\date{}

\begin{document}

\maketitle
\fancyhead{}
\renewcommand{\headrulewidth}{0pt}

\begin{abstract}
Evaluating semantic similarity between videos is a fundamental challenge in computer vision, essential for tasks ranging from out-of-distribution (OOD) detection to video retrieval. However, defining and labeling video similarity is notoriously difficult and expensive due to the complex spatio-temporal nature. In this paper, we propose a novel self-supervised approach that leverages generative uncertainty from text-to-video (T2V) diffusion models to learn semantic similarity without human annotations. Our method is based on the observation that T2V models produce consistent outputs for familiar concepts but exhibit high variance and uncertainty when prompted with specialized concepts. We utilize this behavior to identify stable semantic features within existing pretrained representations, such as VideoMAE and V-JEPA. Specifically, we learn a mask over these embeddings using purely generated data, encouraging the model to retain features that remain consistent across generations of general concepts while discarding those associated with generative noise or uncertainty. Experimental results across three key tasks demonstrate that our learned feature subspaces consistently outperform original pretrained features and baseline feature selection methods.
\end{abstract}

\input{sections/introduction}
\input{sections/related}
\input{sections/method}
\input{sections/experiments}
\input{sections/conclusion}

\section*{Disclosure of AI Assistance}

Generative AI tools assisted with language polishing and with implementation of code used in the reported experiments; all resulting text, code, and experimental outputs were reviewed and verified by the authors. Google Gemini 3.1 Pro was used to generate the specialized text prompts employed in constructing the synthetic training dataset, as described in Appendix~\ref{sec:prompt_gen_supp}. The text-to-video models Wan2.1, CogVideoX1.5-5B, and Mochi 1 were used to generate the video datasets for the proposed method (Section~\ref{sec:syn_data_gen}). Generative AI tools were not used for other tasks requiring disclosure.

\bibliographystyle{conference}
\bibliography{refs}

\clearpage
\appendix
\input{sections/appendix}

\end{document}

%% file: math_commands.tex
\usepackage{amsmath,amsfonts,bm}

\def\eqref#1{equation~\ref{#1}}

\def\1{\bm{1}}

\DeclareMathAlphabet{\mathsfit}{\encodingdefault}{\sfdefault}{m}{sl}
\SetMathAlphabet{\mathsfit}{bold}{\encodingdefault}{\sfdefault}{bx}{n}



%% file: sections/introduction.tex
\section{Introduction}

Learning semantic similarity between videos is important for many tasks, such as out-of-distribution (OOD) detection~\citep{sim2023simple}, video retrieval~\citep{dwibedi2019temporal}, and evaluating generative models~\citep{liu2024evalcrafter}. Yet, learning such similarity from human annotations is challenging: videos contain rich spatio-temporal information, and judging which aspects are semantically relevant is inherently difficult~\citep{zhang2018unreasonable}. Collecting similarity annotations at scale is therefore expensive and often infeasible, particularly for open-domain or generated videos.
This motivates the question: \textit{how can we learn video representations that capture semantic similarity without using labeled data?}

Self-supervised video models such as VideoMAEv2~\citep{wang2023videomaev2} and VJEPA 2.1~\citep{mur2026v} provide strong general-purpose representations across many downstream tasks. However, their training objectives are not designed explicitly for measuring semantic similarity, and their feature spaces can contain dimensions that are irrelevant or even detrimental to similarity-based tasks. As we show experimentally, directly comparing these representations can lead to limited alignment with human judgments and unreliable uncertainty estimates for generated videos.

We investigate whether pretrained video representations can instead be adapted for semantic similarity using only synthetic data from text-to-video (T2V) models. Our key observation is that the diversity of a generative model's outputs provides a natural self-supervised signal.
When a model is given a prompt describing a general concept, it generates videos that tend to be consistent with each other. In contrast, when the prompt describes a specialized concept to the model, the generated videos are more diverse and less consistent (Fig.~\ref{fig:dataset_examples}).
We refer to this diversity as \emph{generative uncertainty} and use it to identify which feature dimensions are most useful for measuring semantic similarity.

To achieve this, we rely entirely on synthetic data generated by a T2V model. We generate videos from two types of prompts: (i) \emph{general} concepts that the model can generate reliably, and (ii) \emph{specialized} concepts (i.e., concepts belonging to highly specific or niche domains) that lead to more uncertain, and therefore diverse outputs.
Particularly, our method does not require generative variability to be a perfect measure of whether a concept was seen during training.
Our hypothesis is weaker: if some representation dimensions remain stable when repeated generations preserve the same semantic content, but become variable when the generated content becomes less consistent, then this difference in variability can reveal dimensions that are useful for semantic similarity.
For each prompt, we generate multiple videos, varying only the initial noise, and learn a mask over pretrained video embeddings, encouraging embeddings of videos generated from the same general prompt to be more similar than videos generated from the same specialized prompt. Importantly, 
all pairs are therefore obtained from videos generated with the \emph{same} prompt, avoiding human similarity annotations or predefined semantic relations between prompts.

We find that the learned feature subspaces consistently outperform the original features and other subspace methods using different T2V models and across three similarity-based tasks: human alignment, OOD detection for generated samples, and video retrieval.
At the same time, the learned masks retain only a fraction of the original feature dimensions, yielding more compact representations. We further evaluate patch-level masking for video object segmentation and observe minimal performance loss despite substantial dimensionality reduction (Appendix~\ref{sec:vos-tracking}).

Overall, our main contributions are as follows:
\begin{itemize}
    \item We identify generative uncertainty in T2V models as a self-supervised signal for semantic similarity: general concepts produce relatively consistent generations, whereas specialized concepts yield more diverse outputs.
    \item Using this signal, we introduce a lightweight, model-agnostic method that learns sparse subspaces of pretrained video representations and improves human-similarity alignment, OOD detection, and video retrieval across different feature encoders.
\end{itemize}

%% file: sections/related.tex
\section{Related work}

\paragraph{Video similarity.}
Defining and quantifying perceptual similarity is a fundamental challenge in computer vision. For static images, deep feature distances have proven very effective~\citep{zhang2018unreasonable} due to their close alignment with human perception. However, extending this paradigm to videos is inherently more complex, as recently shown by ConViS-Bench~\citep{liberatori2025convisbench}. Their benchmark shows that recent powerful models for video understanding, such as V-JEPA 2.1~\citep{mur2026v} and VideoMAE V2~\citep{wang2023videomaev2}, extract representations that align only partially with human judgment. In this work, we build upon these backbones, demonstrating that their raw embeddings can be refined into concentrated subspaces via our targeted masking approach to improve performance in video-similarity-based tasks.

Video similarity learning itself is most commonly explored through the proxy tasks of video retrieval and near-duplicate detection. This is done, for instance, by temporal context aggregation via contrastive learning~\citep{Shao_2021_WACV}, temporal alignment for partial video copy detection~\citep{partial_copy_detection}, and leveraging region attention graphs~\citep{Ng2022VRAGRA}. Other works propose unified frameworks for event retrieval and recounting~\citep{gao2017er3}, suppress irrelevant frames to improve video-to-video retrieval~\citep{jo2024vvs}, or focus explicitly on near-duplicate video retrieval~\citep{he2022learn}. Other works have explored fully self-supervised methods to learn video similarity directly from visual streams~\citep{kordopatis2023s2vs}. \cite{he2022learn} learns near-duplicate retrieval from massive pools of unlabeled videos, while~\cite{kordopatis2023s2vs} relies on synthetic proxy tasks created via heavy data augmentations to learn self-supervised similarity.
However, while these methods are effective for retrieval, they often rely on access to large video datasets. 

One promising avenue, therefore, is to turn to synthetic data. This is largely unexplored in video similarity, although it has seen significant success in other vision domains, such as perceptual image tasks~\citep{wu2023datasetdm} or action recognition~\citep{varol2021synthetic}.
In our work, we use synthetic data to bypass the need for real-world video collections or manual augmentations by using the zero-shot outputs of modern text-to-video models to learn video similarity.

\noindent\textbf{Generative uncertainty.}
Previous works on generative uncertainty largely focus on developing methods to measure the uncertainty of samples produced by the model. For image generation, this is typically done by adapting the original model into a sort of ensemble by introducing randomness and measuring the distance between generations. 
For instance, \cite{berry2024shedding} obtains multiple fine-tuned versions of the label embedding layer and uses the distances between predicted Gaussians to measure uncertainty, while~\cite{jazbec2025generative} draws multiple parameter sets from a Laplace approximation of the final layer and uses the CLIP embedding space to measure distances. Other works make use of the inherent randomness in the diffusion denoising process to obtain uncertainty estimates~\citep{de2025diffusion}. 
In scenarios where text is available, such as a prompt for text-to-image methods, vision-language models can be used to assess similarity to the prompt and measure uncertainty~\citep{franchi2025towards}. 
In this work, we do not propose a new method for measuring generative uncertainty. Rather, to our knowledge, we are the first to exploit this inherent characteristic of generative models to define a self-supervised objective to learn a notion of similarity.

%% file: sections/method.tex
\section{Method}
\begin{figure}[h!]
    \centering
    \includegraphics[width=\linewidth]{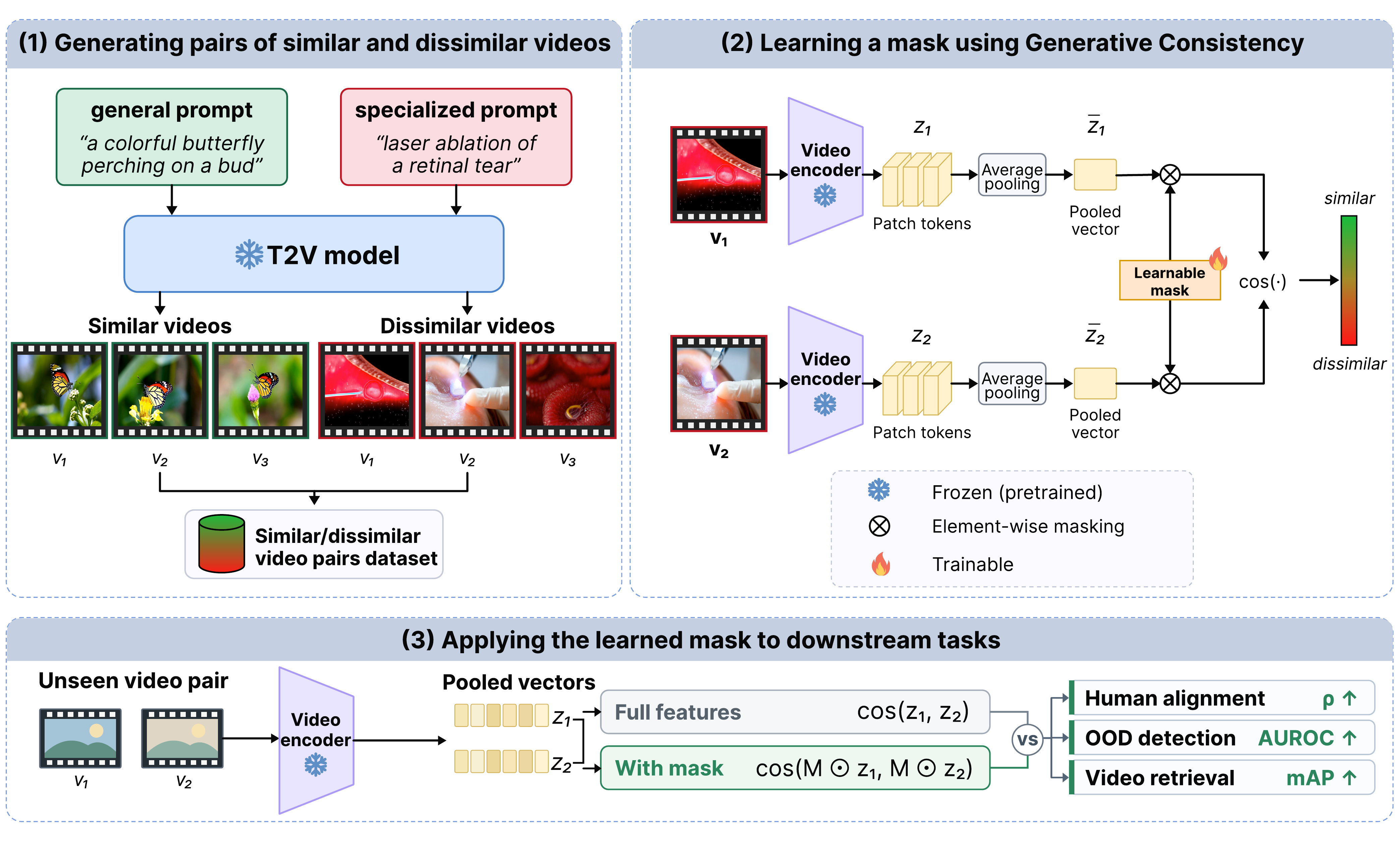}
    \caption{\textbf{Proposed framework overview.} \textbf{(1)} A text-to-video model generates videos from \textit{general} and \textit{specialized} prompts with different random seeds; videos from the same prompt are paired. \textbf{(2)} A frozen video encoder extracts global representations, and a learnable mask is trained to yield high/low cosine similarity for \textit{similar/dissimilar} pairs. \textbf{(3)} For downstream tasks, we mask unseen video representations before cosine similarity. In doing so, our strategy improves alignment with human judgment, out-of-distribution detection, and video retrieval.
}
    \label{fig:method}
\end{figure}

Our method, summarized in Figure~\ref{fig:method}, is driven solely by synthetic data, comprising both the generated videos and their source prompts. Human involvement is limited strictly to the initial domain assumptions and prompt curation, which serves as the foundation for our dataset.

\subsection{Synthetic dataset generation} \label{sec:syn_data_gen}

\textbf{Defining general concepts.} We construct our synthetic dataset using the Wan2.1 (T2V-1.3B) text-to-video model~\citep{wan2025}. A key challenge lies in partitioning the prompt space into \textit{general} and \textit{specialized} concepts relative to the model's underlying training distribution. While the exact training corpus of Wan2.1 is proprietary, it is documented to include broad domains such as natural landscapes, daily human activities, and general scenery. Furthermore, Wan2.1 demonstrates state-of-the-art performance on the VBench benchmark~\citep{huang2023vbench}, which evaluates video generation across a wide array of common semantic categories.
Consequently, we define our \textit{general} set by sampling $N=100$ prompts directly from VBench, representing concepts the model can synthesize with high fidelity and consistency.

\begin{figure*}[t]
    \centering
    \begin{minipage}[t]{0.48\textwidth}
        \centering
        \includegraphics[width=0.8\linewidth]{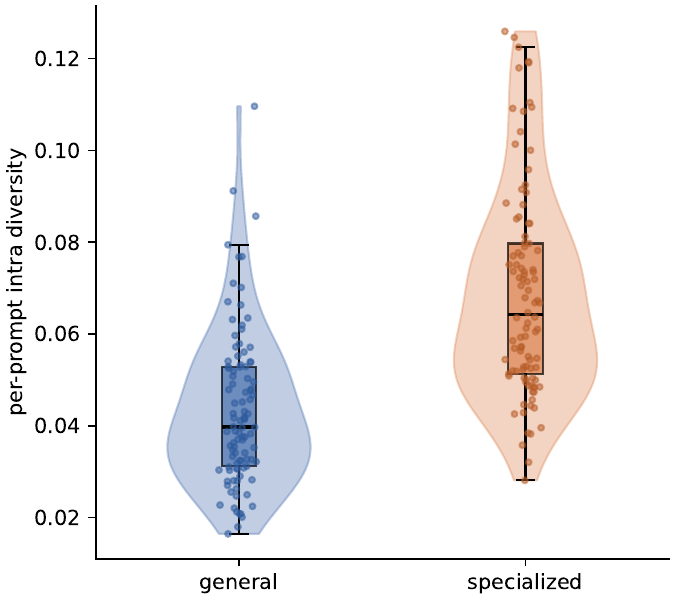}
    \end{minipage}
    \hfill
    \begin{minipage}[t]{0.48\textwidth}
        \centering
        \includegraphics[width=0.8\linewidth]{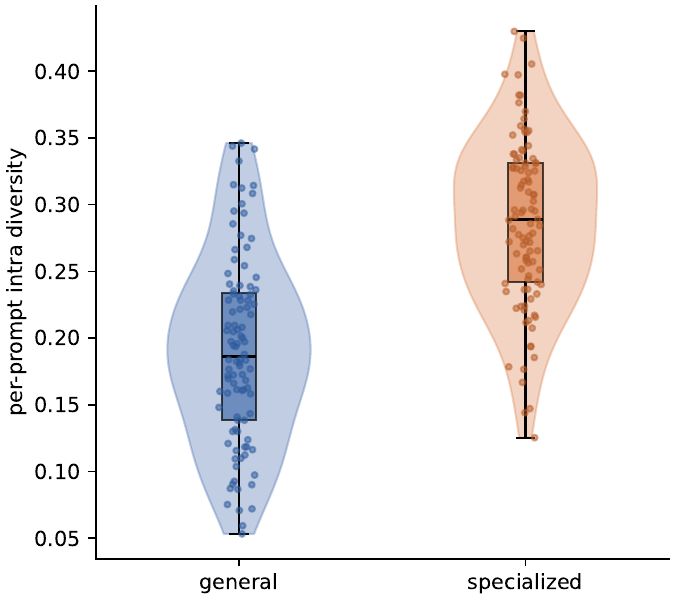}
    \end{minipage}

    \caption{
    \textbf{Per-prompt intra-set feature-space diversity for videos generated by
    Wan2.1, measured using V-JEPA 2.1 Large (left) and VideoMAEv2 Base
    (right).}
    Each point represents one prompt and reports the average pairwise normalized
    cosine dissimilarity among three videos generated with different random seeds.}
    \label{fig:intra_variance_known_unknown_wan}
\end{figure*}

\textbf{Constructing specialized concepts.} Conversely, we define \textit{specialized} prompts as those belonging to highly uncommon or niche domains, such as medical surgery or microscopy, for which current T2V models tend to produce more variable generations in our
experiments~(Fig.\ref{fig:intra_variance_known_unknown_wan}). To construct this set, we leveraged Google's Gemini 3.1 Pro to generate $N=100$ unique prompts from these specialized domains. 
To ensure that our downstream evaluation captures semantic domain shifts rather than superficial linguistic variances, we constrained the language model to match the \textit{general} prompts in terms of length, syntactic complexity, and descriptive granularity. More details in Appendix~\ref{sec:prompt_gen_supp}.

\begin{figure}[t]
    \centering
    \includegraphics[width=0.95\linewidth]{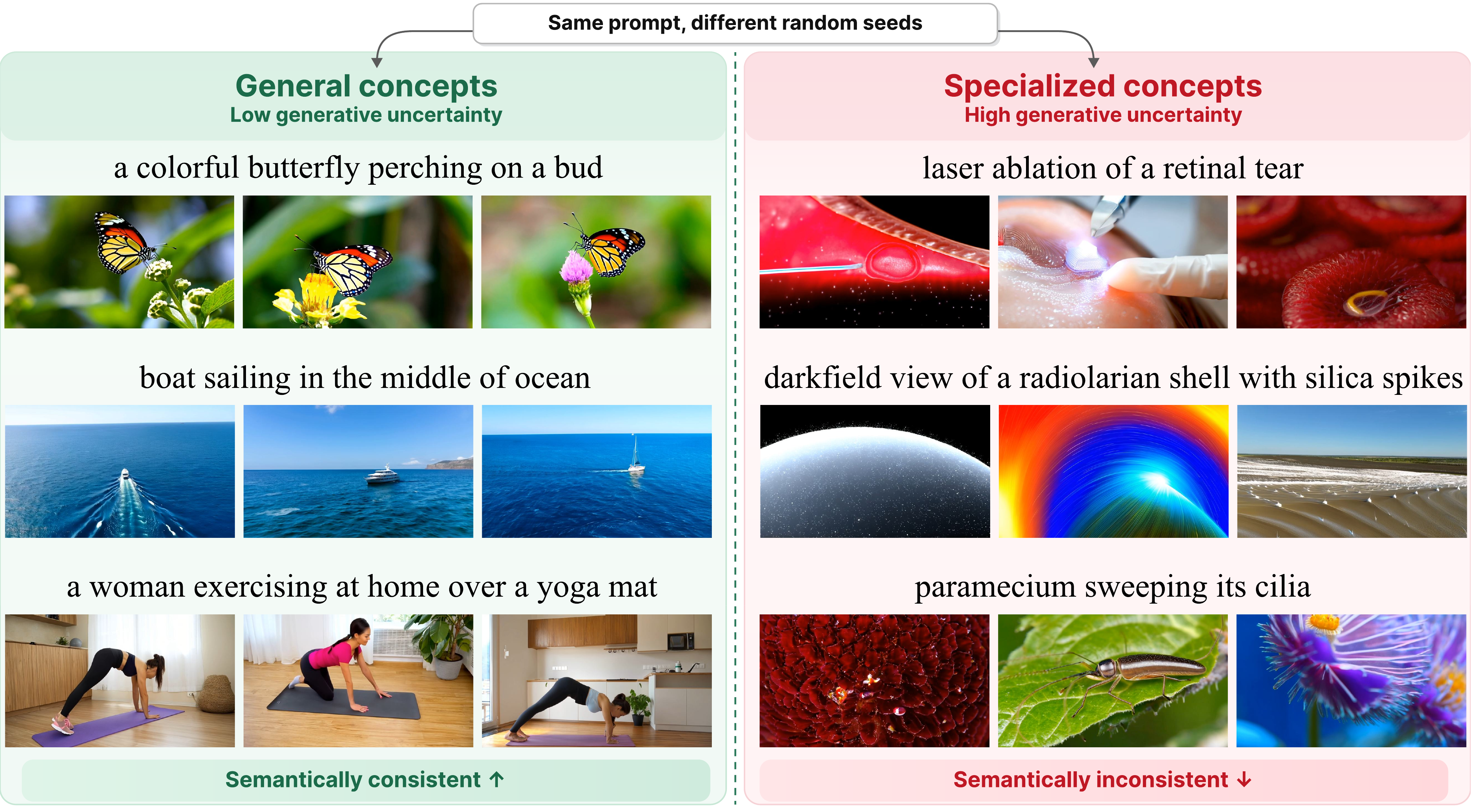}
    \caption{\textbf{Example videos used in our generative consistency objective.} Videos generated by Wan 2.1 with \textit{general} and \textit{specialized} text prompts. Videos from \textit{general} prompts serve as proxies for semantic consistency, while those from \textit{specialized} prompts represent generative divergence.}
    \label{fig:dataset_examples}
\end{figure}

\textbf{Generating the synthetic video dataset.} For each of the 200 total prompts, we synthesized three distinct videos by varying the initial random seed. This process yielded a dataset of \textbf{600} videos. In our framework, videos generated from \textit{general} prompts serve as proxies for semantic consistency (high intra-prompt similarity), while those from \textit{specialized} prompts provide a signal for generative divergence, allowing us to learn a similarity subspace without human supervision. Examples of these synthesized general and specialized concepts are presented in Figure~\ref{fig:dataset_examples} while Figure~\ref{fig:intra_variance_known_unknown_wan} quantitatively shows how general prompts lead to closer embeddings in the chosen feature space.
In Appendix~\ref{sec:additional_video_generators} we present similar findings for other generative models: CogVideoX1.5-5B and Mochi 1.

\subsection{Feature learning via generative consistency} \label{sec:mask_learning}

\textbf{Extracting global video features.} For a given video $v$, we utilize a pretrained ViT-based encoder $\mathcal{E}$ to extract spatio-temporal features. While these architectures produce a sequence of patch tokens of shape $[P, D]$, where $P$ denotes the number of tokens and $D$ the embedding dimension, our objective is to capture global semantic similarity rather than localized cues. Thus, we aggregate tokens into a single global representation $\mathbf{z} = \mathcal{E}(v) \in \mathbb{R}^{D}$ via global average pooling across the token dimension.

\textbf{Learning a soft feature mask.} To identify a subspace that prioritizes stable semantic content, we optimize a learnable soft mask $M \in [0, 1]^D$, which acts as a differentiable feature weighting mechanism. During the training phase, this mask remains continuous to facilitate gradient-based optimization. We utilize the synthetic dataset described in the previous section as a supervisory proxy, optimizing $M$ via the AdamW~\citep{loshchilov2017decoupled} optimizer to minimize the Mean Squared Error (MSE) between the masked cosine similarity of a video pair and a generative stability label $y$. 
Specifically, we set $y=1$ for pairs synthesized from the same \textit{general} prompt (characterizing high semantic consistency) and $y=0$ for those from the same \textit{specialized} prompt (characterizing generative divergence). The training objective is then defined as:

\begin{equation}
s_M(\mathbf{z}_1,\mathbf{z}_2)
=
\frac{
1 + \cos\left(
\mathbf{z}_1 \odot \mathbf{M},
\mathbf{z}_2 \odot \mathbf{M}
\right)
}{2},
\qquad
\mathcal{L}
=
\left(
s_M(\mathbf{z}_1,\mathbf{z}_2) - y
\right)^2,
\quad
y \in \{0,1\}.
\end{equation}

\textbf{From soft weights to sparse feature selection.} To prevent the mask from overfitting to the specific synthetic noise of the T2V model, we employ an early stopping mechanism based on the margin between the average similarities of the general and specialized groups. Once optimization is complete, the mask can be binarized using a threshold of $0.5$ to produce a binary mask $\hat{M} = \mathbb{I}(M > 0.5) \in \{0, 1\}^D$. 
We evaluate both the learned soft mask and its  binarized version for all downstream tasks.

%% file: sections/experiments.tex
\section{Experiments}

We verify the effectiveness of our masked features on three similarity-driven tasks using V-JEPA 2.1 and VideoMAEv2 backbones. To determine whether the uncertainty signal is tied to a specific video generator, we repeat all primary experiments with two additional text-to-video models, CogVideoX1.5-5B~\citep{yang2024cogvideox} and Mochi 1~\citep{genmo2024mochi}. While this section presents results produced with Wan2.1 as the generator, consistent analysis and results for the other two models are provided in Appendix~\ref{app:generator-robustness}. We additionally provide a theoretical interpretation of our method in Appendix~\ref{sec:theory}.

\subsection{Alignment with human perception}
To validate that our learned feature masks select a subspace in which similarity aligns with human judgment, we evaluate our approach on ConViS-Bench \citep{liberatori2025convisbench}. 
This benchmark proposes a dataset of video pairs meant to measure concept-based video similarity. 
Each video pair is annotated by humans with a similarity score across different semantic dimensions (e.g., the main action, subjects, objects, location, and the order of actions). Following ~\cite{liberatori2025convisbench}, we compute cosine similarity between videos in the embedding space of different encoders and measure alignment using Spearman's $\rho$ and Kendall's $\tau$ rank correlation coefficients, evaluating how well a model's latent representations align with human judgment across fine-grained semantic dimensions.

\textbf{Learned masks improve human similarity alignment.} As shown in Table \ref{tab:convisbench}, applying our learned masks yields a consistent improvement in human alignment across the evaluated models. For instance, with V-JEPA 2.1-L, both the discrete binary mask and the continuous soft mask strictly outperform the unmasked baseline across all five conceptual categories. We observe notable increases in the $\rho$ correlation for "Main Subjects" (+6.0 for the binary mask, +6.1 for the soft mask) and "Main Objects", with the soft mask achieving the best overall performance. A similar trend of uniform improvement across all categories is evident for VideoMAEv2-B, where binary and soft mask perform similarly. These results indicate that our training objective successfully forces the models to prioritize features that align with human judgment. 
These significant gains in human alignment can be achieved with high feature sparsity when using the binary mask: the V-JEPA 2.1-L and VideoMAEv2-B models achieve superior semantic alignment using only 32.2\% and 42.0\% of their original embedding dimensions, respectively.
This demonstrates that our learning approach
effectively isolates a concentrated subspace of features that aligns with human perception. 
In contrast, unsupervised feature-selection baselines generally do not provide comparable gains. Detailed comparison to various baselines and per-category results are provided in Appendix~\ref{app:convisbench}.

\begin{table}[t]
\centering
\caption{\textbf{Alignment with human judgement on ConViS-Bench.}
Spearman's $\rho$ and Kendall's $\tau$ correlations with human judgement ($\times100$).
\emph{Sp.} denotes the sparsity of the learned mask; soft masks are dense by construction.
Our masked features better match the human notion of global video similarity.
Best results are \textbf{bolded}, second-best \underline{underlined}.}
\label{tab:convisbench}

\scriptsize
\renewcommand{\arraystretch}{1.25}
\setlength{\tabcolsep}{3.5pt}

\settowidth{\encw}{\textbf{VideoMAEv2-B}}
\addtolength{\encw}{10pt}
\addtolength{\encw}{2\tabcolsep}
\setlength{\encblk}{\ht\strutbox}
\addtolength{\encblk}{\dp\strutbox}
\setlength{\encblk}{3.75\encblk}

\resizebox{\textwidth}{!}{%
\begin{tabular}{
  c
  >{\columncolor{StubMeth}}l
  >{\columncolor{SpBody}}c
  >{\columncolor{GrpABody}}c >{\columncolor{GrpABody}}c
  >{\columncolor{GrpBBody}}c >{\columncolor{GrpBBody}}c
  >{\columncolor{GrpABody}}c >{\columncolor{GrpABody}}c
  >{\columncolor{GrpBBody}}c >{\columncolor{GrpBBody}}c
  >{\columncolor{GrpABody}}c >{\columncolor{GrpABody}}c
  >{\columncolor{AvgBody}}c  >{\columncolor{AvgBody}}c}
\toprule

& & \cellcolor{SpHead}
& \multicolumn{2}{>{\columncolor{GrpAHead}}c}{\textbf{Main Action}}
& \multicolumn{2}{>{\columncolor{GrpBHead}}c}{\textbf{Main Subjects}}
& \multicolumn{2}{>{\columncolor{GrpAHead}}c}{\textbf{Main Objects}}
& \multicolumn{2}{>{\columncolor{GrpBHead}}c}{\textbf{Location}}
& \multicolumn{2}{>{\columncolor{GrpAHead}}c}{\textbf{Actions Order}}
& \multicolumn{2}{>{\columncolor{AvgHead}}c}{\textbf{Average}} \\

\textbf{Encoder}
& \textbf{Method}
& \cellcolor{SpHead}\textbf{Sp.\,(\%)}
& \cellcolor{GrpAMet}$\rho$ & \cellcolor{GrpAMet}$\tau$
& \cellcolor{GrpBMet}$\rho$ & \cellcolor{GrpBMet}$\tau$
& \cellcolor{GrpAMet}$\rho$ & \cellcolor{GrpAMet}$\tau$
& \cellcolor{GrpBMet}$\rho$ & \cellcolor{GrpBMet}$\tau$
& \cellcolor{GrpAMet}$\rho$ & \cellcolor{GrpAMet}$\tau$
& \cellcolor{AvgMet}$\rho$  & \cellcolor{AvgMet}$\tau$ \\

\midrule

& Baseline (no mask)
& \na
& 24.0 & 16.30 & 33.4 & 22.84 & 24.7 & 16.83
& 49.3 & 35.04 & 24.3 & 16.65 & 31.14 & 21.53 \\

& Binary mask
& 67.8
& \underline{26.9} & \underline{18.25} & \underline{39.4} & \underline{27.14}
& \underline{30.3} & \underline{20.57} & \underline{50.5} & \underline{35.74}
& \underline{30.1} & \underline{20.75} & \underline{35.44} & \underline{24.49} \\

\multirow{-3}{*}{\enc{V-JEPA 2.1-L}}
& Soft mask
& \na
& \textbf{27.7} & \textbf{18.78} & \textbf{39.5} & \textbf{27.24}
& \textbf{30.8} & \textbf{20.89} & \textbf{51.3} & \textbf{36.41}
& \textbf{30.9} & \textbf{21.23} & \textbf{36.04} & \textbf{24.91} \\

\midrule

& Baseline (no mask)
& \na
& 47.0 & 33.02 & 44.6 & 31.46 & 42.4 & 29.96 & 45.8 & 32.26 & 44.7 & 31.37 & 44.90 & 31.61 \\

& Binary mask
& 58.0
& \underline{47.9} & \underline{33.62} & \textbf{46.3} & \textbf{32.63} & \textbf{44.1} & \textbf{30.98} & \textbf{46.9} & \textbf{33.09} & \underline{45.3} & \underline{31.67} & \textbf{46.10} & \textbf{32.40} \\

\multirow{-3}{*}{\enc{VideoMAEv2-B}}
& Soft mask
& \na
& \textbf{48.0} & \textbf{33.74} & \underline{46.1} & \underline{32.47} & \underline{44.0} & \underline{30.97} & \underline{46.7} & \underline{32.92} & \textbf{45.4} & \textbf{31.79} & \underline{46.04} & \underline{32.38} \\

\bottomrule
\end{tabular}%
}
\end{table}

\subsection{Video generation OOD detection}\label{sec:i2v_ood}
\textbf{Transfer to generative OOD detection.}
We focus on OOD detection for image-to-video generation, where a single input image is used as context to generate future frames. When this context image lies far from the distribution expected by the generative model, the generated video may become unreliable, unstable, or semantically inconsistent. Detecting such OOD inputs is therefore important for estimating when future-frame generation can be trusted. Since OOD detection for generative models often relies on measuring similarity between multiple generated outputs, this experiment tests whether a mask learned from synthetic text-to-video uncertainty can also improve similarity-based uncertainty estimation in another generative task.

\textbf{Evaluation setup.} 
We employ NVIDIA Cosmos-Predict2~\citep{nvidia2025cosmospredict2} as our base Image-to-Video (I2V) generative model. To establish a controlled ID baseline, we perform fine-tuning on the BDD100k dataset ~\citep{bdd100k}; consequently, videos from the BDD100k validation set are treated as ID samples. For OOD evaluation, we select four diverse datasets: UVEB (underwater footage)~\citep{xie2024uveb}, MedVidBench (medical surgery)~\citep{su2026medgrpo}, HAM10000 (dermatoscopic images)~\citep{tschandl_2018_the}, and WikiArt (artistic images)\footnote{\url{https://huggingface.co/datasets/huggan/wikiart}}. While the exact training corpus of Cosmos is proprietary, its design focus on real-world physics and robotics provides high confidence that these four datasets represent significant distributional shifts.
To quantify OOD scores, we adopt two procedures inspired by \cite{franchi2025towards}:
\begin{itemize}
    \item \textbf{2XDM}~\citep{berry2024shedding, franchi2025towards}: We generate two distinct videos from the same context image using different initial random noise. Then, both videos are mapped to an embedding space via a video encoder, and the cosine dissimilarity between their embeddings serves as the OOD score. 
    \item \textbf{DDPM-OOD}~\citep{Graham_2023_CVPR, franchi2025towards}: We first generate a base video $v_0$ conditioned on a context image using $K=35$ diffusion steps. We then apply $k \in [1, K)$ steps of the forward diffusion process to this video and repeat the denoising steps. This procedure is performed three times with $k \in \{10, 20, 30\}$ to generate three new videos, denoted as $v_1$, $v_2$, and $v_3$, respectively. The final score is computed as the average of the cosine dissimilarities between $v_0$ and each of the three generated videos ($v_1$, $v_2$, $v_3$) in the embedding space. 
\end{itemize}

\textbf{Masked features improve OOD detection.}
This evaluation pipeline is conducted independently for each OOD dataset and repeated with both detection methods (2XDM, DDPM-OOD) using our learned feature mask. As shown in Table~\ref{tab:OOD-results}, applying the feature mask yields consistent improvements across all datasets, feature spaces, and detection methods. The binary masks show the highest gains, with improvements in AUROC up to $34\%$ and FPR@95 up to $51\%$, while retaining a fraction of the feature dimensions.
We again compare our learned mask against alternative masking strategies and unsupervised feature-selection methods, as well as training with a standard similarity objective, which we further discuss in the ablation studies and Table~\ref{tab:ood-vs-sim_full}.

\begin{table}[t]
\centering
\caption{\textbf{Video generation OOD detection across four diverse datasets.}
Videos are generated using Cosmos-Predict2, with BDD100k as the in-distribution (ID) baseline.
Applying binary and soft feature masks consistently improves OOD detection performance.
Best results are \textbf{bolded} and second-best results are \underline{underlined}.}
\label{tab:OOD-results}

\scriptsize
\renewcommand{\arraystretch}{1.2}
\setlength{\tabcolsep}{3.5pt}

\resizebox{\textwidth}{!}{%
\begin{tabular}{l l
  >{\columncolor{XDMBody}}c >{\columncolor{XDMBody}}c >{\columncolor{XDMBody}}c
  >{\columncolor{DDPMBody}}c >{\columncolor{DDPMBody}}c >{\columncolor{DDPMBody}[\tabcolsep][0pt]}c
  >{\columncolor{XDMBody}[0pt][\tabcolsep]}c >{\columncolor{XDMBody}}c >{\columncolor{XDMBody}}c
  >{\columncolor{DDPMBody}}c >{\columncolor{DDPMBody}}c >{\columncolor{DDPMBody}}c}
\toprule

& &
\multicolumn{6}{>{\columncolor{EncoderBlue}[\tabcolsep][0pt]}c}
{\textcolor{white}{\textbf{V-JEPA 2.1-L}}} &
\multicolumn{6}{>{\columncolor{EncoderBlue}[0pt][\tabcolsep]}c}
{\textcolor{white}{\textbf{VideoMAEv2-B}}} \\[1pt]

& &
\multicolumn{3}{>{\columncolor{XDMHead}}c}{\textbf{2XDM}} &
\multicolumn{3}{>{\columncolor{DDPMHead}[\tabcolsep][0pt]}c}{\textbf{DDPM-OOD}} &
\multicolumn{3}{>{\columncolor{XDMHead}[0pt][\tabcolsep]}c}{\textbf{2XDM}} &
\multicolumn{3}{>{\columncolor{DDPMHead}}c}{\textbf{DDPM-OOD}} \\

\multirow{-3}{*}{\textbf{OOD Dataset}} &
\multirow{-3}{*}{\textbf{Method}} &
\cellcolor{XDMMetric}AUROC\,$\uparrow$ &
\cellcolor{XDMMetric}AUPR\,$\uparrow$ &
\cellcolor{XDMMetric}FPR@95\,$\downarrow$ &
\cellcolor{DDPMMetric}AUROC\,$\uparrow$ &
\cellcolor{DDPMMetric}AUPR\,$\uparrow$ &
\multicolumn{1}{>{\columncolor{DDPMMetric}[\tabcolsep][0pt]}c}
{FPR@95\,$\downarrow$} &
\multicolumn{1}{>{\columncolor{XDMMetric}[0pt][\tabcolsep]}c}
{AUROC\,$\uparrow$} &
\cellcolor{XDMMetric}AUPR\,$\uparrow$ &
\cellcolor{XDMMetric}FPR@95\,$\downarrow$ &
\cellcolor{DDPMMetric}AUROC\,$\uparrow$ &
\cellcolor{DDPMMetric}AUPR\,$\uparrow$ &
\cellcolor{DDPMMetric}FPR@95\,$\downarrow$ \\

\midrule

\multirow{3}{*}{\textbf{UVEB}}
& Unmasked
& 78.42 & 73.77 & 57.20
& 76.37 & 70.31 & \underline{63.60}
& 71.08 & 68.50 & 77.60
& 75.50 & 73.33 & \textbf{83.20} \\

& Binary mask
& \textbf{86.77} & \textbf{85.72} & \textbf{51.20}
& \textbf{89.52} & \textbf{88.04} & \textbf{42.80}
& \textbf{72.31} & \textbf{69.27} & \underline{73.60}
& \textbf{77.47} & \textbf{75.10} & \underline{84.80} \\

& Soft mask
& \underline{85.79} & \underline{84.48} & \underline{52.00}
& \underline{89.11} & \underline{87.35} & \textbf{42.80}
& \underline{72.21} & \underline{68.90} & \textbf{73.20}
& \underline{77.37} & \underline{74.81} & \textbf{83.20} \\

\midrule

\multirow{3}{*}{\textbf{MedVidBench}}
& Unmasked
& 74.66 & 74.18 & 76.00
& 68.11 & 68.47 & 86.00
& 78.59 & 80.73 & 82.40
& 77.01 & 79.46 & \textbf{79.60} \\

& Binary mask
& \textbf{88.80} & \textbf{89.65} & \textbf{61.20}
& \textbf{91.45} & \textbf{92.40} & \textbf{48.40}
& \textbf{80.43} & \textbf{82.08} & \textbf{79.60}
& \textbf{79.32} & \textbf{81.58} & 83.20 \\

& Soft mask
& \underline{88.09} & \underline{89.00} & \underline{62.80}
& \underline{91.15} & \underline{91.97} & \underline{49.20}
& \underline{80.14} & \underline{81.49} & \underline{80.00}
& \underline{78.71} & \underline{80.91} & \underline{81.20} \\

\midrule

\multirow{3}{*}{\textbf{HAM10000}}
& Unmasked
& 70.80 & 61.21 & 63.20
& 87.67 & 87.95 & 51.60
& 68.38 & 62.42 & 78.40
& 86.23 & 88.37 & \textbf{67.60} \\

& Binary mask
& \textbf{80.09} & \textbf{74.44} & \textbf{55.60}
& \textbf{95.57} & \textbf{96.03} & \textbf{25.20}
& \textbf{70.16} & \textbf{63.64} & \underline{75.20}
& \textbf{88.02} & \textbf{89.87} & 71.60 \\

& Soft mask
& \underline{79.03} & \underline{72.74} & \underline{58.00}
& \underline{95.40} & \underline{95.87} & \underline{26.00}
& \underline{69.95} & \underline{63.35} & \textbf{74.80}
& \underline{87.96} & \underline{89.72} & \underline{68.80} \\

\midrule

\multirow{3}{*}{\textbf{WikiArt}}
& Unmasked
& 63.10 & 55.82 & 65.20
& 69.80 & 65.03 & 82.40
& 57.78 & 51.05 & 82.00
& 63.88 & 63.16 & \underline{89.20} \\

& Binary mask
& \textbf{71.25} & \textbf{66.30} & \textbf{64.80}
& \textbf{82.06} & \textbf{78.11} & \textbf{66.00}
& \textbf{58.89} & \textbf{51.52} & \underline{79.20}
& \textbf{65.95} & \textbf{64.37} & 90.00 \\

& Soft mask
& \underline{69.88} & \underline{64.56} & \underline{64.80}
& \underline{81.53} & \underline{77.27} & \underline{68.00}
& \underline{58.60} & \underline{51.28} & \textbf{77.60}
& \underline{65.55} & \underline{63.59} & \textbf{88.40} \\

\bottomrule
\end{tabular}%
}
\end{table}

\subsection{Video retrieval}
Measuring semantic similarity between high-dimensional video embeddings is fundamental for video retrieval. We focus on UCF101~\citep{soomro2012ucf101} and HMDB51~\citep{kuehne2011hmdb}, where given a query video, the objective is to retrieve videos belonging to the same class as the query, thus showing high semantic similarity. We thus benchmark our masked features on both datasets using both V-JEPA 2.1 and VideoMAEv2 feature spaces.

\textbf{Masked features improve video retrieval.}
Table~\ref{tab:retrieval_wan} presents the mAP results for the standard features and the soft and binary masked ones. Our results indicate that our learned masks consistently improve retrieval performance on both datasets and features spaces, with a slight decrease for VideoMAEv2 on HMDB51. Additional results with different video generators are reported in Appendix~\ref{app:generator-robustness} Table~\ref{tab:retrieval_mochi_cogvideox}, with the strongest performance coming from masks trained on CogVideoX.

\newcommand{\blockrow}[1]{\rowcolor{hdr!10}\multicolumn{7}{l}{%
  \hspace{-2pt}\textcolor{hdr}{\bfseries #1}}\\}

\begin{table}[t]
  \centering
  \caption{Class-level \textbf{video retrieval mAP (\%)} on UCF101 and HMDB51.
    Best results are \textbf{bolded} and second-best results are \underline{underlined}.}
  \label{tab:retrieval_wan}
  \setlength{\tabcolsep}{7pt}
  \renewcommand{\arraystretch}{1.22}
  \resizebox{0.6\textwidth}{!}{%
  \begin{tabular}{l l ccc}
    \arrayrulecolor{hdr}\specialrule{1.1pt}{0pt}{0pt}
    \rowcolor{hdr}
    \color{white}\bfseries\gp Dataset & \cellcolor{hdr}\color{white}\bfseries Encoder &
    \multicolumn{3}{c}{\cellcolor{hdr}\color{white}\bfseries Retrieval mAP (\%)} \\
    \arrayrulecolor{white}\cmidrule(lr){3-5}\arrayrulecolor{hdr}
    \rowcolor{hdr}
    \cellcolor{hdr} & \cellcolor{hdr} &
    \color{white}Unmasked & \color{white}Binary mask & \color{white}Soft mask \\
    \specialrule{1.1pt}{0pt}{0pt}\arrayrulecolor{black}

    \multirow{2}{*}{\dset{UCF101}}
      & V-JEPA 2.1 Large & 40.82 & \runr{41.34} & \best{42.37} \\
      & VideoMAEv2-Base  & 95.52 & \runr{95.61} & \best{95.65} \\
    \arrayrulecolor{black!25}\midrule\arrayrulecolor{black}
    \multirow{2}{*}{\dset{HMDB51}}
      & V-JEPA 2.1 Large & 17.84 & \runr{18.40} & \best{18.99} \\
      & VideoMAEv2-Base  & \best{46.94} & 46.39 & \runr{46.66} \\
    \arrayrulecolor{hdr}\specialrule{1.1pt}{2pt}{0pt}\arrayrulecolor{black}
  \end{tabular}}
\end{table}

\subsection{Ablations}
To validate the design choices of our proposed feature masking approach, we conduct extensive ablation studies. We aim to answer three primary questions: \textbf{(Q1)} Is our \emph{generative consistency} objective necessary, or would a standard similarity-based objective suffice? \textbf{(Q2)} Given the high sparsity of our learned masks, is the performance gain simply a byproduct of dimensionality reduction, or are we identifying a highly specific subset of features? And \textbf{(Q3)} Do our highly sparse masks destroy essential visual and structural information? 

\begin{table}[t]
\centering
\caption{\textbf{Ablation study on feature masking strategies and training objectives.}
Results are averaged across four OOD datasets (UVEB, MedVidBench, HAM10000, and WikiArt).
We compare our masking strategy against the standard similarity objective (\textbf{Q1}) and
unsupervised feature-selection baselines (\textbf{Q2}) evaluated at equal sparsity
($k=329$ for V-JEPA 2.1-L, $k=322$ for VideoMAEv2-B).
}
\label{tab:ood-vs-sim_full}

\scriptsize
\renewcommand{\arraystretch}{1.2}
\setlength{\tabcolsep}{3.5pt}

\resizebox{\textwidth}{!}{%
\begin{tabular}{l
  >{\columncolor{XDMBody}}c >{\columncolor{XDMBody}}c >{\columncolor{XDMBody}}c
  >{\columncolor{DDPMBody}}c >{\columncolor{DDPMBody}}c >{\columncolor{DDPMBody}[\tabcolsep][0pt]}c
  >{\columncolor{XDMBody}[0pt][\tabcolsep]}c >{\columncolor{XDMBody}}c >{\columncolor{XDMBody}}c
  >{\columncolor{DDPMBody}}c >{\columncolor{DDPMBody}}c >{\columncolor{DDPMBody}}c}
\toprule

&
\multicolumn{6}{>{\columncolor{EncoderBlue}[\tabcolsep][0pt]}c}
{\textcolor{white}{\textbf{V-JEPA 2.1-L}}} &
\multicolumn{6}{>{\columncolor{EncoderBlue}[0pt][\tabcolsep]}c}
{\textcolor{white}{\textbf{VideoMAEv2-B}}} \\[1pt]

&
\multicolumn{3}{>{\columncolor{XDMHead}}c}{\textbf{2XDM}} &
\multicolumn{3}{>{\columncolor{DDPMHead}[\tabcolsep][0pt]}c}{\textbf{DDPM-OOD}} &
\multicolumn{3}{>{\columncolor{XDMHead}[0pt][\tabcolsep]}c}{\textbf{2XDM}} &
\multicolumn{3}{>{\columncolor{DDPMHead}}c}{\textbf{DDPM-OOD}} \\

\multirow{-3}{*}{\textbf{Method}} &
\cellcolor{XDMMetric}AUROC\,$\uparrow$ &
\cellcolor{XDMMetric}AUPR\,$\uparrow$ &
\cellcolor{XDMMetric}FPR@95\,$\downarrow$ &
\cellcolor{DDPMMetric}AUROC\,$\uparrow$ &
\cellcolor{DDPMMetric}AUPR\,$\uparrow$ &
\multicolumn{1}{>{\columncolor{DDPMMetric}[\tabcolsep][0pt]}c}
{FPR@95\,$\downarrow$} &
\multicolumn{1}{>{\columncolor{XDMMetric}[0pt][\tabcolsep]}c}
{AUROC\,$\uparrow$} &
\cellcolor{XDMMetric}AUPR\,$\uparrow$ &
\cellcolor{XDMMetric}FPR@95\,$\downarrow$ &
\cellcolor{DDPMMetric}AUROC\,$\uparrow$ &
\cellcolor{DDPMMetric}AUPR\,$\uparrow$ &
\cellcolor{DDPMMetric}FPR@95\,$\downarrow$ \\

\midrule

Baseline (unmasked)
& 71.75 & 66.25 & 65.40
& 75.49 & 72.94 & 70.90
& 68.96 & 65.68 & 80.10
& 75.66 & 76.08 & 79.90 \\

Similarity objective
& \underline{74.39} & \underline{70.30} & 65.50
& \underline{80.97} & \underline{78.30} & 67.90
& 68.90 & 65.81 & 82.80
& 76.06 & \underline{77.04} & 81.50 \\

\textbf{Ours (binary mask)}
& \textbf{81.73} & \textbf{79.03} & \textbf{58.20}
& \textbf{89.65} & \textbf{88.65} & \textbf{45.60}
& \textbf{70.45} & \underline{66.63} & \textbf{76.90}
& \textbf{77.69} & \textbf{77.73} & 82.40 \\

\midrule

PCA projection
& 70.33 & 65.08 & 69.80
& 70.52 & 67.91 & 77.50
& 63.07 & 59.18 & 87.80
& 67.89 & 66.36 & 81.20 \\

Random mask
& 70.91 & 65.62 & 66.95
& 72.68 & 70.54 & 74.30
& 67.84 & 63.80 & 81.03
& 74.73 & 74.59 & 80.70 \\

Self-attention
& 70.29 & 64.78 & \underline{64.80}
& 69.68 & 67.63 & 81.10
& \underline{70.03} & \textbf{67.02} & 80.50
& \underline{76.18} & 76.57 & \underline{78.60} \\

Variance (top-$k$)
& 70.27 & 64.80 & 65.40
& 69.61 & 67.50 & 80.90
& 69.49 & 66.39 & 80.50
& 75.46 & 75.86 & \underline{78.60} \\

PCA loading
& 70.22 & 64.72 & 65.90
& 69.70 & 67.56 & 80.30
& 69.54 & 66.48 & 82.30
& 75.49 & 75.82 & \textbf{77.40} \\

Laplacian score
& 69.14 & 63.74 & 69.50
& 67.54 & 66.11 & 83.50
& 69.03 & 65.80 & 80.60
& 75.13 & 75.13 & 81.00 \\

Correlation filter
& 73.20 & 68.04 & 65.10
& 78.89 & 76.92 & \underline{65.60}
& 69.30 & 66.38 & \underline{79.30}
& 75.67 & 76.18 & 80.60 \\

\bottomrule
\end{tabular}%
}
\end{table}

\textbf{The necessity of generative consistency~(Q1).}
A naive alternative for learning the feature mask is to optimize for a standard similarity objective. Under this setup, we follow the same training strategy but fundamentally change the definition of positive (similar) and negative (dissimilar) pairs to learn our mask.
In particular, we ignore the idea behind general and specialized concepts and label a pair of videos as semantically similar ($y=1$) if generated from the same prompt and build negative pairs ($y=0$) using videos that come from different prompts. Once again, given that we have generated our dataset with a T2V model, there is no need for manual labeling.

As shown in Table \ref{tab:ood-vs-sim_full}, while this similarity objective (row 2) provides a marginal improvement over the baseline (unmasked) embeddings, falling significantly short of our proposed method. 
This suggests that video pairs labeled based on generative consistency, i.e., belonging to \textit{general} or \textit{specialized} concepts, capture more significant semantic differences compared to just considering two videos similar because they have the same description (source prompt).   

\textbf{Learned masks vs dimensionality reduction~(Q2).}
Our learned binary masks exhibit high sparsity. For instance, with V-JEPA 2.1-L, the mask zeros out 67.8\% of the features using a threshold $\theta=0.5$, effectively utilizing only $329$ of the original $1024$ embedding dimensions. Similar patterns emerge for VideoMAEv2 (58.0\% sparsity). To ensure this improvement is not an artifact of simply reducing the feature space, we compare our method against several dimensionality reduction and feature selection baselines, including random masking, variance-based feature selection, PCA-based methods, Laplacian Score, correlation-based filtering, and a self-attention-derived mask, which we further describe in the Appendix~\ref{app:baselines}.
The results in the second half of Tab.~\ref{tab:ood-vs-sim_full} validate our approach. 
Across both models and OOD methods, none of the other unsupervised feature selection or dimensionality reduction techniques perform as well as the subspace identified by our learning strategy.
This indicates that the features most critical for similarity are not trivial to define.   
Notably, our method drastically reduces the False Positive Rate (FPR@95). For instance, in the V-JEPA DDPM-OOD setup, our method reduces the FPR@95 from the unmasked baseline of 70.90\% down to 45.60\%. 
These results confirm that these models possess highly capable, latent subspaces for OOD detection, but that they must be isolated using a targeted objective rather than generic dimensionality reduction.
Additional results in Table~\ref{tab:convisbench_full} (Appendix~\ref{app:convisbench}) confirm this trend on ConViS-Bench: our generative-consistency masks outperform all the aforementioned baselines at matched sparsity.

\subsubsection{Preservation of visual information}
To verify that our highly sparse masks do not destroy essential visual information (\textbf{Q3}), we evaluate the learned representations both quantitatively and qualitatively.

\begin{figure}[h!]
    \centering
    \includegraphics[width=0.95\linewidth]{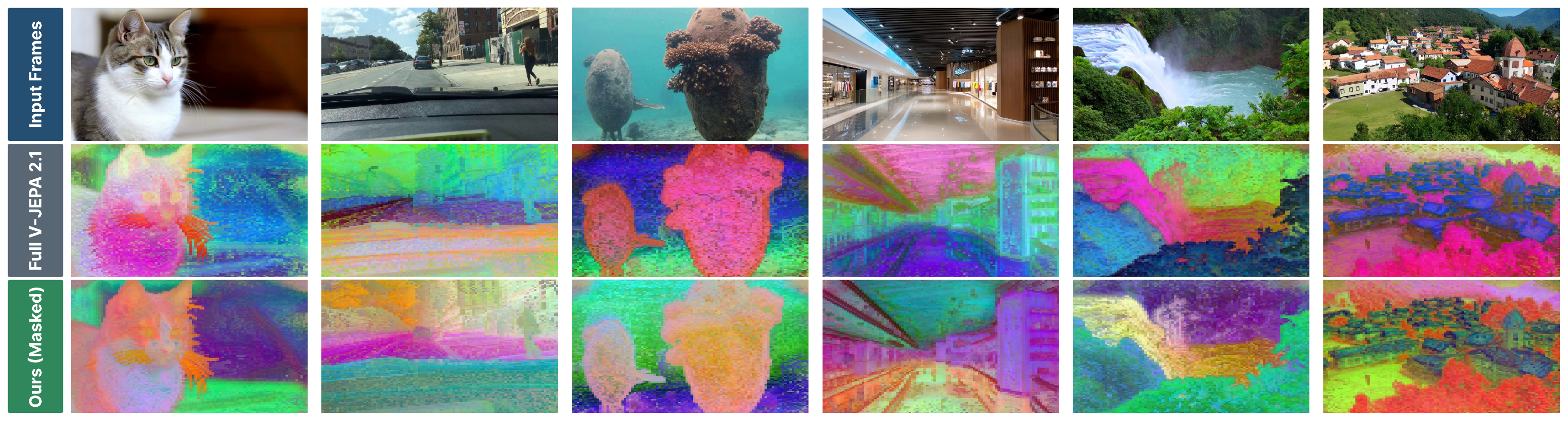}
    \caption{\textbf{PCA visualization of V-JEPA 2.1-L patch-level features mapped to RGB channels.} (Top) Original input frames. (Middle) Full V-JEPA features. (Bottom) Our masked representations. 
    }
    \label{fig:pca_viz}
\end{figure}

\vspace{1ex}\noindent\textbf{PCA feature visualization.}
To qualitatively illustrate this preserved spatial structure, we visualize the feature space directly using PCA. For a given video frame, we extract patch-level features using the V-JEPA 2.1-L encoder, compute PCA to extract the top three principal components, and map them directly to RGB color channels. The visual comparisons are presented in Figure \ref{fig:pca_viz}, displaying the original input frames (top row), the PCA visualizations of the unmasked features (middle row), and our masked representations (bottom row). The masked features exhibit remarkably similar structural outlines, object boundaries, and background separation compared to the full baseline, demonstrating their retention of core visual features despite the high sparsity.

\textbf{Patch clustering for spatial reconstruction.} 
Following prior work on analyzing the spatial structure encoded in visual representations \citep{DHANACHANDRA2015764}, we cluster both masked and unmasked patch features of one frame per video to obtain a piecewise segmentation, map the segments to their mean pixel value, and compute the $\operatorname{MSE}_k$ to the original image, where $k$ denotes the number of clusters. 
We report the final $\operatorname{MSE}_k$ averaged across all evaluated frames in the BDD100k dataset across varying cluster counts showing that even after masking out a substantial portion of the embedding dimensions, the representations still retain much of the original visual and structural information. Details in Appendix~\ref{app:reconstruction}.
\begin{figure}[h!]
    \centering
    \includegraphics[width=0.5\linewidth]{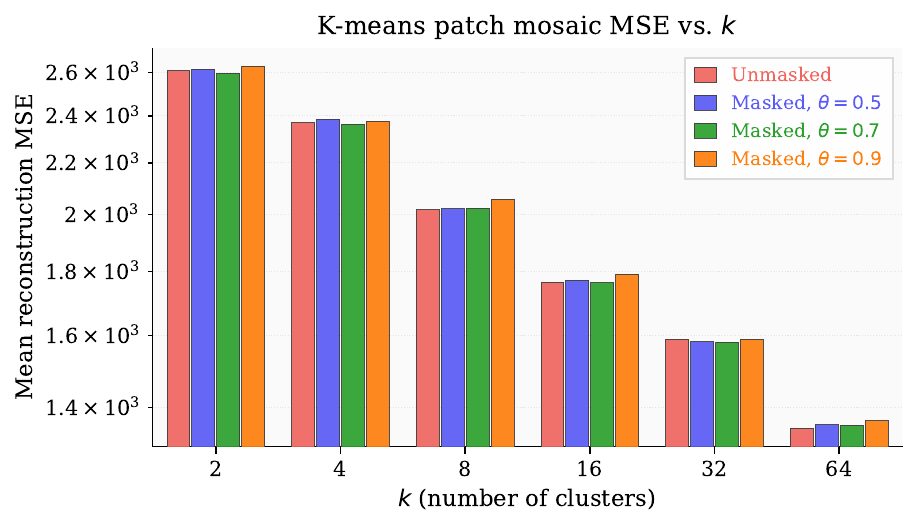}
    \caption{\textbf{Quantitative information loss evaluation via K-means mosaic reconstruction.} The plot compares the Mean Squared Error (MSE) between the original RGB frames and the mosaic reconstructions across varying cluster counts ($k$). The V-JEPA 2.1-L masked representations evaluate multiple binarization thresholds ($\theta$).}
    \label{fig:kmeans_mosaic_threshold}
\end{figure}

\vspace{-0.5cm}

%% file: sections/conclusion.tex
\section{Conclusion}
In this paper, we introduced a self-supervised framework for learning semantic video similarity by leveraging generative uncertainty in text-to-video diffusion models. Rather than relying on human similarity annotations or predefined similarity metrics, our approach uses the consistency of generated videos as a supervisory signal: general concepts tend to produce stable generations, while specialized concepts induce higher generative divergence. We exploit this behavior to learn sparse masks over pretrained video representations, isolating feature subspaces that better capture semantic similarity. Across multiple backbones and tasks, the learned masked representations consistently improve alignment with human similarity judgments, enhance out-of-distribution detection for generated videos, and maintain or improve video retrieval performance. Importantly, these gains are achieved while discarding a large fraction of the original embedding dimensions, leading to more compact and efficient representations. Our dense-task experiments further show that the learned global masks preserve meaningful local structure, enabling efficient patch-level matching with minimal loss in segmentation quality.
Finally, we reproduce the experiments across three different text-to-video models, validating our results on independent sets of videos.
These results suggest that generative models can provide more than synthetic data: their uncertainty can serve as a useful signal for learning semantically useful subspaces. 

%% file: sections/appendix.tex
\renewcommand{\thefigure}{\thesection.\arabic{figure}}
\renewcommand{\thetable}{\thesection.\arabic{table}}
\renewcommand{\theequation}{\thesection.\arabic{equation}}

\makeatletter
\@addtoreset{figure}{section}
\@addtoreset{table}{section}
\@addtoreset{equation}{section}
\makeatother

\section{Theoretical Analysis of Generative Variability}
\label{sec:theory}

\subsection{Why is generative variability a useful learning signal?}
\label{sec:theory_variability}

Our method relies on a simple observation: when the same text prompt is
sampled several times, the generated videos can be either very consistent or
highly variable. We use this variability as a self-supervised signal to identify
which dimensions of a pretrained video representation are useful for semantic
similarity.

Importantly, our analysis does not assume that a specialized prompt is
necessarily absent from the training set of the video generator. It also does
not require all generative variability to be epistemic uncertainty.
Instead, we only study the variability of repeated generations conditioned on
the same prompt, and show why differences in this variability can reveal useful
feature dimensions.

\paragraph{Variability of repeated video generations.}

Let $p$ denote a text prompt and let
\begin{equation}
    \mathbf{v} \sim p_{\theta}(\cdot \mid p)
\end{equation}
be a video generated by a stochastic text-to-video model.
For a fixed prompt $p$, we consider two independent generations
\begin{equation}
    \mathbf{v},\mathbf{v}'
    \stackrel{\mathrm{i.i.d.}}{\sim}
    p_{\theta}(\cdot \mid p).
    \label{eq:iid_video_generations}
\end{equation}

Intuitively, if the generator represents the content of $p$ in a stable way,
$V$ and $V'$ should contain similar semantic information.
If several different semantic realizations are plausible, repeated generations
can instead be substantially different.

To formalize this idea, suppose that the possible videos generated from $p$
can be grouped into $K_p$ semantic outcomes,
\begin{equation}
    \mathcal{V}_{p}
    =
    \bigcup_{k=1}^{K_p}
    \mathcal{C}_{p,k},
\end{equation}
where videos in the same set $\mathcal{C}_{p,k}$ express the same main
semantic content.
Let
\begin{equation}
    g_p(\mathbf{v}) \in \{1,\ldots,K_p\}
\end{equation}
denote the semantic outcome of $\mathbf{v}$, and define
\begin{equation}
    \pi_k(p)
    =
    \Pr\left(g_p(\mathbf{v})=k\mid p\right).
\end{equation}

We can then define the semantic variability of the video generator as
\begin{equation}
    \mathcal{V}_{\mathrm{sem}}(p)
    =
    \Pr\left(
        g_p(\mathbf{v}) \neq g_p(\mathbf{v}')
        \mid p
    \right).
    \label{eq:video_semantic_variability}
\end{equation}

\paragraph{Proposition 1 (Pairwise disagreement measures semantic variability).}
For two independent generations from the same prompt,
\begin{equation}
    \boxed{
    \mathcal{V}_{\mathrm{sem}}(p)
    =
    1-\sum_{k=1}^{K_p}\pi_k(p)^2.
    }
    \label{eq:video_collision}
\end{equation}

\textit{Proof.}
Since $V$ and $V'$ are independent conditioned on $p$,
\begin{align}
    \Pr\left(g_p(\mathbf{v})=g_p(\mathbf{v}')\mid p\right)
    &=
    \sum_{k=1}^{K_p}
    \Pr\left(g_p(\mathbf{v})=k\mid p\right)
    \Pr\left(g_p(\mathbf{v}')=k\mid p\right)
    \\
    &=
    \sum_{k=1}^{K_p}\pi_k(p)^2.
\end{align}
Taking the complementary event gives
~\eqref{eq:video_collision}.
\hfill$\square$

Equation~\ref{eq:video_collision} gives a simple interpretation of repeated
generation. If most of the probability mass is concentrated on one semantic
outcome, two generations are likely to agree and
$\mathcal{V}_{\mathrm{sem}}(p)$ is small.
If the probability mass is spread over several possible outcomes,
$\mathcal{V}_{\mathrm{sem}}(p)$ becomes larger.

As above, let $g_p(\mathbf{v})$ denote the semantic content of
$\mathbf{v}$. We define the semantic variability of the generator as
\begin{equation}
    \mathcal{V}_{\mathrm{sem}}(p)
    =
    \Pr\left(
        g_p(\mathbf{v})
        \neq
        g_p(\mathbf{v}')
        \mid p
    \right).
    \label{eq:semantic_variability_similarity}
\end{equation}

We now define the ideal semantic similarity between two videos as
\begin{equation}
    s^\star(\mathbf{v},\mathbf{v}')
    =
    \mathbf{1}
    \left[
        g_p(\mathbf{v})
        =
        g_p(\mathbf{v}')
    \right].
    \label{eq:ideal_semantic_similarity}
\end{equation}

Thus, $s^\star=1$ when the two videos have the same semantic content and
$s^\star=0$ otherwise.

\paragraph{Proposition 2 (Uncertainty and semantic similarity).}
For two independent videos generated from the same prompt,
\begin{equation}
    \boxed{
    \mathbb{E}
    \left[
        s^\star(\mathbf{v},\mathbf{v}')
        \mid p
    \right]
    =
    1-\mathcal{V}_{\mathrm{sem}}(p).
    }
    \label{eq:uncertainty_similarity}
\end{equation}

\textit{Proof.}
By definition,
\begin{align}
    \mathbb{E}
    \left[
        s^\star(\mathbf{v},\mathbf{v}')
        \mid p
    \right]
    &=
    \Pr
    \left(
        g_p(\mathbf{v})
        =
        g_p(\mathbf{v}')
        \mid p
    \right)
    \\
    &=
    1-
    \Pr
    \left(
        g_p(\mathbf{v})
        \neq
        g_p(\mathbf{v}')
        \mid p
    \right)
    \\
    &=
    1-\mathcal{V}_{\mathrm{sem}}(p).
\end{align}

\paragraph{Interpretation.}
Proposition~\ref{eq:uncertainty_similarity} gives a direct connection between
generative uncertainty and semantic similarity.
If the uncertainty of the generator is low, two independent generations are
likely to have the same semantic content, and their expected semantic
similarity is high.
Conversely, if the uncertainty is high, two generations are more likely to
have different semantic contents, and their expected semantic similarity is
lower.  We want to insist on the assumption of the ideal semantic similarity of~ \eqref{eq:ideal_semantic_similarity}, which is linked with the fact that two videos are similar if they have similar concepts. Under this assumption, uncertainty can be linked to the ideal semantic similarity and provide a natural supervision signal for learning video similarity, In our work, we exploit this uncertainty signal to learn similarity over the feature space defined by a pretrained video encoder $\mathcal{E}$.

\section{Implementation details}
\subsection{\textit{Specialized} prompt generation} \label{sec:prompt_gen_supp}
We ensure that the \textit{specialized} prompts differ from the \textit{general} prompts strictly in their semantic domain, rather than through superficial linguistic artifacts (e.g., varying sequence lengths, different grammatical structures, or stylistic shifts). To achieve this controlled domain shift, we leveraged Google's Gemini 3.1 Pro to generate a parallel dataset of 100 specialized concepts.

We explicitly constrained the language model to synthesize prompts focused on highly specialized, non-everyday domains (specifically, clinical surgery and cellular microscopy) while perfectly mirroring the structural distribution of the general prompts. The exact prompt provided to the language model was as follows:

\begin{quote}
\textit{``I have a dataset of 100 general concept prompts. I need an equivalent set of 100 specialized concept prompts that perfectly match the general dataset in detail, tone, exact length distribution, and concise structural style. The specialized prompts must focus exclusively on the domains of clinical surgery (50 prompts) and cellular microscopy (50 prompts). Additionally, they must be entirely devoid of everyday objects, animals, and human subjects (e.g., 'scissors', 'dog', 'person').''}

\texttt{`general\_prompts.txt' attached}
\end{quote}

By enforcing these constraints, we reduce the likelihood that the observed generative divergence is driven by superficial linguistic differences between the general and specialized prompts.

Here are some examples of \textit{general} and \textit{specialized} prompts used for video generation:

\definecolor{knownbg}{RGB}{238,248,242}
\definecolor{knownborder}{RGB}{65,145,95}
\definecolor{unknownbg}{RGB}{252,241,241}
\definecolor{unknownborder}{RGB}{190,80,80}

\medskip

\noindent
\begin{minipage}[t]{0.485\linewidth}
\begin{tcolorbox}[
    colback=knownbg,
    colframe=knownborder,
    boxrule=0.8pt,
    arc=2mm,
    title=\textbf{General concepts},
    coltitle=white,
    colbacktitle=knownborder,
    fonttitle=\small,
    left=2mm,
    right=2mm,
    top=1.5mm,
    bottom=1.5mm,
    equal height group=prompts
]
\small
\begin{itemize}[leftmargin=4mm,itemsep=3pt,topsep=1pt]
    \item ``cheetah lying on the grass''
    \item ``scenery of desert landscape''
    \item ``a relaxing scenery of beach view under cloudy sky''
    \item ``a chef holding and checking kitchen utensils''
    \item ``people packing their furniture''
\end{itemize}
\end{tcolorbox}
\end{minipage}
\hfill
\begin{minipage}[t]{0.485\linewidth}
\begin{tcolorbox}[
    colback=unknownbg,
    colframe=unknownborder,
    boxrule=0.8pt,
    arc=2mm,
    title=\textbf{Specialized concepts},
    coltitle=white,
    colbacktitle=unknownborder,
    fonttitle=\small,
    left=2mm,
    right=2mm,
    top=1.5mm,
    bottom=1.5mm,
    equal height group=prompts
]
\small
\begin{itemize}[leftmargin=4mm,itemsep=3pt,topsep=1pt]
    \item ``skin closure with sutures''
    \item ``surgical retractor holding back thick adipose tissue''
    \item ``macrophage engulfing rod bacteria''
    \item ``fluorescent view of actin filaments contracting muscle''
    \item ``sealing a bleeding vessel with bipolar cautery''
\end{itemize}
\end{tcolorbox}
\end{minipage}

\medskip

Representative prompts are listed above, and additional qualitative examples of generated videos from the general and specialized domains are included later in this appendix.
\subsection{Learnable mask training}

We optimize the continuous soft mask by partitioning our synthetic dataset of general and specialized video pairs into training (80\%) and validation (20\%) sets. The mask parameters are updated using the AdamW optimizer with a learning rate of $1 \times 10^{-3}$ and a batch size of 128. To prevent overfitting, we employ an early stopping mechanism on the validation set. Features from the used video encoder were extracted and stored before training to speed up the process.
Given that we use $y=1$ for positive pairs and $y=0$ for negative pairs, during training the computed cosine similarity is rescaled from $[-1,1]$ to $[0,1]$ range to match the target $y$.
All training procedures were executed on a single NVIDIA H100 GPU, with an average training time of approximately 1 minute.
\subsection{Cosmos-Predict2 finetuning}\label{sec:cosmos_finetune}
To ensure that driving scenarios from the BDD100k dataset could indeed be considered as In-Distribution (ID) for the Cosmos model, we performed a short fine-tuning.
For our experiments, we used \texttt{cosmos-predict2} 2B model, and fine-tuned it on BDD100k with batch size 32 and learning rate $1.1 \times 10^{-6}$ for 5000 iterations, approximately 12 hours using 16 NVIDIA H100 GPUs.
The resulting model was then used to perform all the image-to-video generation needed for the experiment as described in sec.~\ref{sec:i2v_ood}.

\section{Additional analysis}

\subsection{Additional video generators}
\label{sec:additional_video_generators}

\begin{figure*}[t]
    \centering

    \begin{subfigure}[t]{0.48\textwidth}
        \centering
        \includegraphics[width=\linewidth]{
            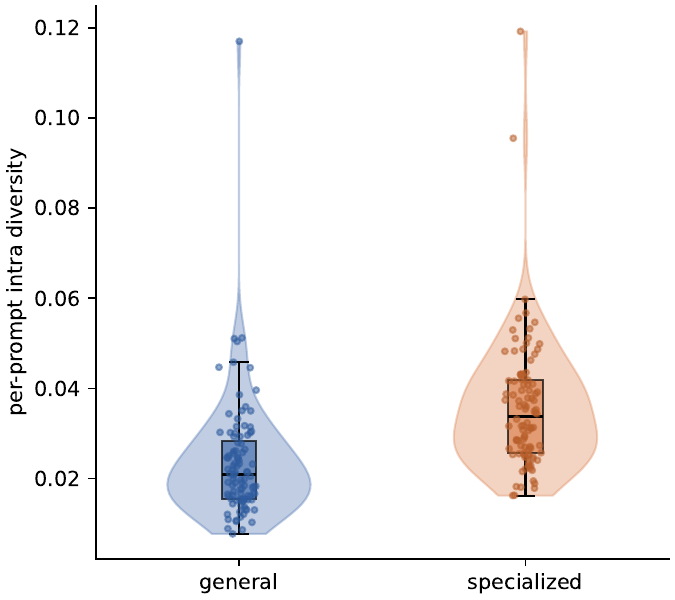
        }
        \caption{CogVideoX1.5-5B -- V-JEPA 2.1 Large} 
        \label{fig:intra_variance_cogvideox_vjepa}
    \end{subfigure}
    \hfill
    \begin{subfigure}[t]{0.48\textwidth}
        \centering
        \includegraphics[width=\linewidth]{
            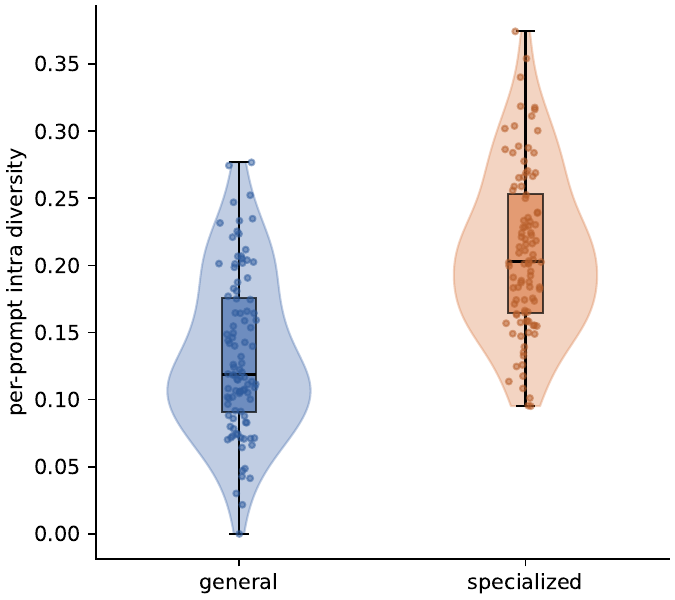
        }
        \caption{CogVideoX1.5-5B -- VideoMAEv2 Base}
        \label{fig:intra_variance_cogvideox_videomae}
    \end{subfigure}

    \vspace{0.5em}

    \begin{subfigure}[t]{0.48\textwidth}
        \centering
        \includegraphics[width=\linewidth]{
            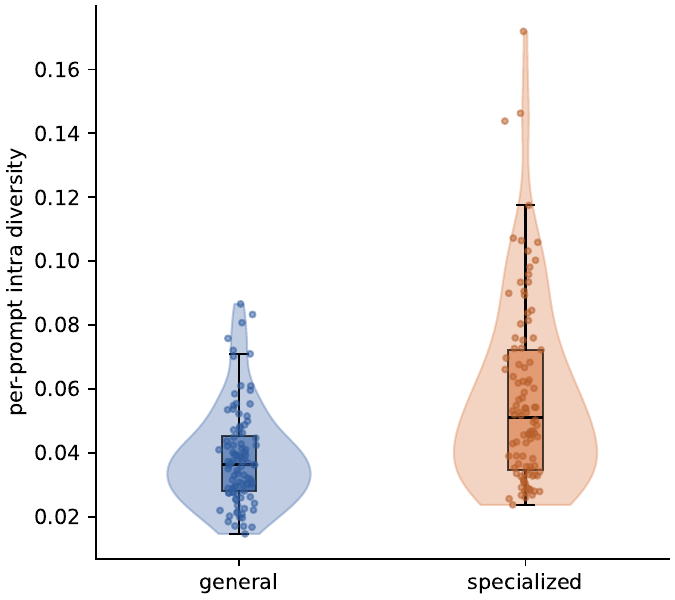
        }
        \caption{Mochi 1 -- V-JEPA 2.1 Large}
        \label{fig:intra_variance_mochi_vjepa}
    \end{subfigure}
    \hfill
    \begin{subfigure}[t]{0.48\textwidth}
        \centering
        \includegraphics[width=\linewidth]{
            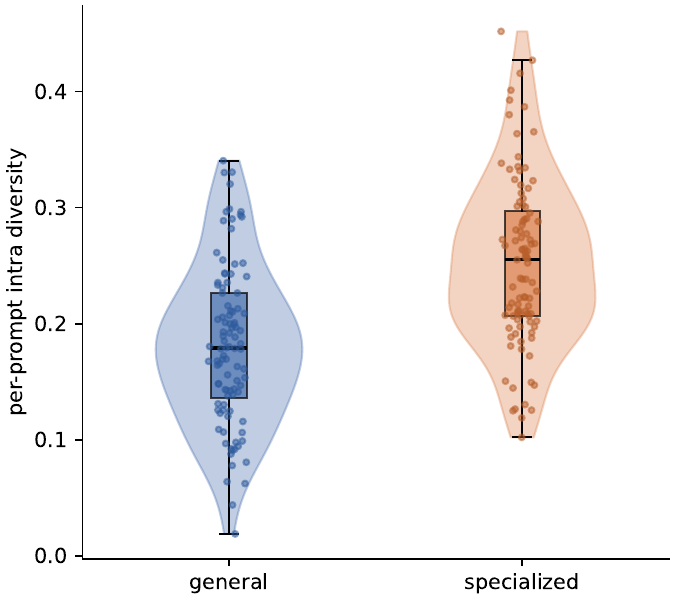
        }
        \caption{Mochi 1 -- VideoMAEv2 Base}
        \label{fig:intra_variance_mochi_videomae}
    \end{subfigure}

    \caption{
        \textbf{Per-prompt intra-set feature-space diversity for videos generated by
        CogVideoX1.5-5B (top row) and Mochi 1 (bottom row), measured using
        V-JEPA 2.1 Large (left column) and VideoMAEv2 Base (right column).}
        Each point represents one prompt and reports the average pairwise normalized
        cosine dissimilarity among three videos generated with different random seeds.
        Specialized prompts exhibit higher intra-prompt diversity than general prompts
        in both feature spaces, indicating lower consistency across generations.
    }
    \label{fig:intra_variance_known_unknown}
\end{figure*}

We repeat the intra-prompt diversity analysis using videos generated by
CogVideoX1.5-5B and Mochi 1. Figure~\ref{fig:intra_variance_known_unknown} shows the results for CogVideoX1.5-5B
and Mochi 1. Consistent with the Wan2.1 results in the main paper,
specialized prompts exhibit higher intra-prompt feature-space diversity than general
prompts in both V-JEPA 2.1-L and VideoMAEv2-B feature spaces. Together with the
downstream results obtained using separately learned Wan2.1-, CogVideoX-, and
Mochi-based masks, these results show that the proposed learning signal is not
specific to a particular video diffusion generator.

\begin{figure*}[t]
    \centering
    \includegraphics[
        width=\textwidth
    ]{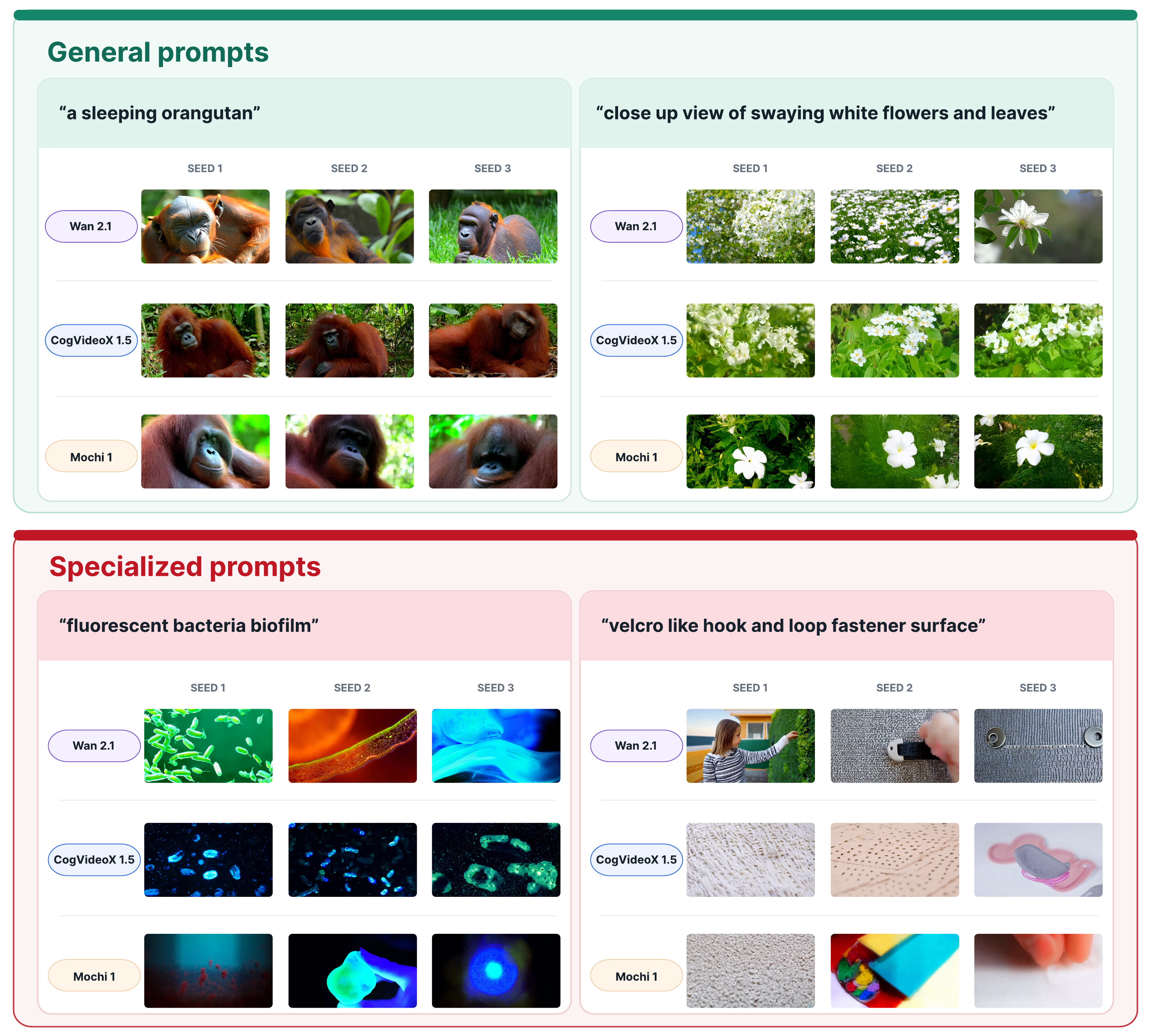}
    \caption{\textbf{Cross-generator qualitative comparison of intra-prompt
    consistency.}
    For each general and specialized prompt, we show three videos generated
    with different random seeds using Wan~2.1, CogVideoX1.5, and Mochi~1.
    Each tile contains sampled frames from one generated video.
    General prompts produce comparatively consistent generations across seeds,
    whereas specialized prompts exhibit greater visual and semantic variation
    across all three generators.}
    \label{fig:cross_generator_examples}
\end{figure*}

We further provide qualitative examples from three different text-to-video generators.
Figure~\ref{fig:cross_generator_examples} compares generations obtained from
general and specialized prompts using three random seeds. Across all three
generators, general prompts tend to produce more consistent content across
seeds, while specialized prompts exhibit greater visual and semantic variation.
This suggests that the proposed generative uncertainty signal is not specific
to a particular video generator.

\subsection{Feature selection and Dimensionality reduction Baselines}
\label{app:baselines}

We compare our generative-consistency mask with several unsupervised
feature-selection and dimensionality-reduction methods.
The goal is to test whether the gains of our method can be explained simply
by reducing the representation size, keeping high-variance features, removing
redundant dimensions, or preserving the geometry of the original embedding
space.
For all masking baselines, we retain the same number of feature dimensions
as our learned binary mask: $k=329/1024$ for V-JEPA 2.1-L and $k=322/768$ for
VideoMAEv2-B.

\paragraph{Random mask.}
We randomly select $k$ feature dimensions from the original representation.
This serves as a sparsity control and tests whether the improvements arise
simply from reducing the representation dimensionality.

\paragraph{Variance (top-$k$).}
We rank feature dimensions by their empirical variance across videos and
retain the $k$ highest-variance dimensions. This baseline favors
statistically active channels but does not explicitly account for semantic
relevance.

\paragraph{PCA loading.}
We perform PCA ~\citep{jolliffe2005principal} on the video representations and rank the original feature
dimensions according to their contribution to the dominant principal
components. The top-$k$ dimensions are retained, allowing us to test whether
our mask primarily captures the dominant variance structure of the
pretrained representation.

\paragraph{Laplacian score.}
We use Laplacian Score ~\citep{NIPS2005_b5b03f06} as an unsupervised feature-selection method that
favors dimensions that preserve the local neighborhood structure of the
embedding space. Features with the lowest scores are retained. This
baseline evaluates whether preserving the representation's local manifold
structure is sufficient to obtain similar improvements.

\paragraph{Correlation filter.}
We rank features according to their redundancy with other channels using
pairwise feature correlations. Dimensions that provide comparatively
non-redundant information are preferred. ~\citep{hall2000correlation} This baseline tests whether the
benefits of our method can be explained by removing redundant channels.

\paragraph{Self-attention mask.}
We additionally construct a feature mask from the encoder's final-block
self-attention activations. Since self-attention is defined over tokens
rather than embedding dimensions, we combine token-level attention weights
with the corresponding patch features to obtain a channel-importance score,
average these scores over the generated videos, and retain the top-$k$
channels.

This baseline tests whether the encoder's internal attention saliency alone
identifies a subspace suitable for semantic similarity. As shown in
Tables~\ref{tab:convisbench_full} and~\ref{tab:ood-vs-sim_full}, the
attention-derived mask does not provide improvements comparable to our
generative-consistency objective.

\subsection{Full Results on ConViS-Bench}
\label{app:convisbench}

Table~\ref{tab:convisbench_full} reports the complete results on
ConViS-Bench across the five semantic dimensions: main action, main subjects,
main objects, location, and action order.
We compare the original pretrained representations with the
feature-selection baselines introduced in
Appendix~\ref{app:baselines}, as well as our binary and soft masks.
For all sparse methods, we keep exactly the same number of dimensions as our
binary mask. Therefore, the comparison isolates the effect of
\emph{which} dimensions are selected, rather than the effect of sparsity
itself.

\paragraph{Results with V-JEPA 2.1-L.}
For V-JEPA 2.1-L, generic feature-selection methods do not improve the
alignment with human judgements.
Random selection, variance-based selection, PCA loading, Laplacian Score,
correlation filtering, and the self-attention mask all obtain lower average
correlations than the original representation.
For instance, the unmasked features obtain an average Spearman correlation of
$31.14$, while the strongest non-learned baseline reaches $29.74$.

In contrast, our binary mask increases the average Spearman correlation to
$35.44$ and the average Kendall correlation from $21.53$ to $24.49$.
The improvement is also consistent across all five semantic dimensions.
This result is obtained while discarding $67.8\%$ of the original feature
dimensions.
The soft mask performs slightly better, reaching an average Spearman
correlation of $36.04$.
This suggests that the generative-consistency objective identifies dimensions
that are particularly useful for semantic similarity, rather than simply
removing low-variance or redundant features.

\paragraph{Results with VideoMAEv2-B.}
The same general trend is observed with VideoMAEv2-B, although the gains are
smaller.
Unlike V-JEPA, some generic feature-selection methods can slightly improve the
unmasked representation.
For example, the random mask increases the average Spearman correlation from
$44.90$ to $45.42$.
This suggests that the VideoMAEv2 representation contains some redundancy and
that reducing its dimensionality can already be beneficial. However, our learned masks still provide the strongest overall alignment.

\begin{table}[t]
\centering
\caption{\textbf{Alignment with human judgement on ConViS-Bench.}
Spearman's $\rho$ and Kendall's $\tau$ correlations with human judgement ($\times100$).
\emph{Sp.} denotes the sparsity of the mask; soft masks are dense by construction.
Unsupervised feature selectors are fit on the video embeddings alone and matched to the
sparsity of our learned mask.
Best results are \textbf{bolded}, second-best \underline{underlined}.}
\label{tab:convisbench_full}

\scriptsize
\renewcommand{\arraystretch}{1.25}
\setlength{\tabcolsep}{3.5pt}

\settowidth{\encw}{\textbf{VideoMAEv2-B}}
\addtolength{\encw}{10pt}
\addtolength{\encw}{2\tabcolsep}
\setlength{\encblk}{\ht\strutbox}
\addtolength{\encblk}{\dp\strutbox}
\setlength{\encblk}{11.25\encblk}

\resizebox{\textwidth}{!}{%
\begin{tabular}{
  c
  >{\columncolor{StubMeth}}l
  >{\columncolor{SpBody}}c
  >{\columncolor{GrpABody}}c >{\columncolor{GrpABody}}c
  >{\columncolor{GrpBBody}}c >{\columncolor{GrpBBody}}c
  >{\columncolor{GrpABody}}c >{\columncolor{GrpABody}}c
  >{\columncolor{GrpBBody}}c >{\columncolor{GrpBBody}}c
  >{\columncolor{GrpABody}}c >{\columncolor{GrpABody}}c
  >{\columncolor{AvgBody}}c  >{\columncolor{AvgBody}}c}
\toprule

& & \cellcolor{SpHead}
& \multicolumn{2}{>{\columncolor{GrpAHead}}c}{\textbf{Main Action}}
& \multicolumn{2}{>{\columncolor{GrpBHead}}c}{\textbf{Main Subjects}}
& \multicolumn{2}{>{\columncolor{GrpAHead}}c}{\textbf{Main Objects}}
& \multicolumn{2}{>{\columncolor{GrpBHead}}c}{\textbf{Location}}
& \multicolumn{2}{>{\columncolor{GrpAHead}}c}{\textbf{Actions Order}}
& \multicolumn{2}{>{\columncolor{AvgHead}}c}{\textbf{Average}} \\

\textbf{Encoder}
& \textbf{Method}
& \cellcolor{SpHead}\textbf{Sp.\,(\%)}
& \cellcolor{GrpAMet}$\rho$ & \cellcolor{GrpAMet}$\tau$
& \cellcolor{GrpBMet}$\rho$ & \cellcolor{GrpBMet}$\tau$
& \cellcolor{GrpAMet}$\rho$ & \cellcolor{GrpAMet}$\tau$
& \cellcolor{GrpBMet}$\rho$ & \cellcolor{GrpBMet}$\tau$
& \cellcolor{GrpAMet}$\rho$ & \cellcolor{GrpAMet}$\tau$
& \cellcolor{AvgMet}$\rho$  & \cellcolor{AvgMet}$\tau$ \\

\midrule

& Baseline (no mask)
& \na
& 24.0 & 16.30 & 33.4 & 22.84 & 24.7 & 16.83 & 49.3 & 35.04 & 24.3 & 16.65 & 31.14 & 21.53 \\
& Random mask
& 67.8
& 22.8 & 15.42 & 31.8 & 21.77 & 23.3 & 15.84 & 47.3 & 33.53 & 22.8 & 15.61 & 29.60 & 20.43 \\
& Variance (top-$k$)
& 67.8
& 21.7 & 14.67 & 31.4 & 21.51 & 21.3 & 14.51 & 47.8 & 33.75 & 22.1 & 15.16 & 28.86 & 19.92 \\
& PCA loading
& 67.8
& 21.5 & 14.54 & 31.3 & 21.45 & 21.1 & 14.43 & 47.8 & 33.76 & 21.9 & 15.10 & 28.72 & 19.86 \\
& Laplacian score
& 67.8
& 21.7 & 14.66 & 30.9 & 21.16 & 20.7 & 14.14 & 47.0 & 33.26 & 22.0 & 15.10 & 28.46 & 19.66 \\
& Correlation filter
& 67.8
& 23.9 & 16.19 & 31.1 & 21.34 & 23.0 & 15.55 & 46.2 & 32.66 & 24.5 & 16.84 & 29.74 & 20.52 \\
& Self-attention mask
& 67.8
& 21.6 & 14.54 & 31.1 & 21.23 & 21.0 & 14.32
& 47.6 & 33.67 & 21.8 & 14.94 & 28.62 & 19.74 \\
& Binary mask (Ours)
& 67.8
& \underline{26.9} & \underline{18.25} & \underline{39.4} & \underline{27.14} & \underline{30.3} & \underline{20.57} & \underline{50.5} & \underline{35.74} & \underline{30.1} & \underline{20.75} & \underline{35.44} & \underline{24.49} \\
\multirow{-9}{*}{\enc{V-JEPA 2.1-L}}
& Soft mask (Ours)
& \na
& \textbf{27.7} & \textbf{18.78} & \textbf{39.5} & \textbf{27.24} & \textbf{30.8} & \textbf{20.89} & \textbf{51.3} & \textbf{36.41} & \textbf{30.9} & \textbf{21.23} & \textbf{36.04} & \textbf{24.91} \\

\midrule

& Baseline (no mask)
& \na
& 47.0 & 33.02 & 44.6 & 31.46 & 42.4 & 29.96 & 45.8 & 32.26 & 44.7 & 31.37 & 44.90 & 31.61 \\
& Random mask
& 58.0
& 47.3 & 33.31 & 45.6 & 32.11 & 43.0 & 30.29 & 46.3 & 32.68 & 44.9 & 31.47 & 45.42 & 31.97 \\
& Variance (top-$k$)
& 58.0
& 45.9 & 32.16 & 43.0 & 30.09 & 40.4 & 28.43 & 44.4 & 31.33 & 43.9 & 30.69 & 43.52 & 30.54 \\
& PCA loading
& 58.0
& 45.8 & 32.13 & 43.1 & 30.19 & 40.4 & 28.44 & 44.7 & 31.47 & 43.7 & 30.60 & 43.54 & 30.57 \\
& Laplacian score
& 58.0
& 46.4 & 32.55 & 45.6 & 32.17 & 42.6 & 30.05 & 45.9 & 32.45 & 44.6 & 31.23 & 45.02 & 31.69 \\
& Correlation filter
& 58.0
& 45.1 & 31.70 & 42.6 & 29.83 & 40.2 & 28.28 & 42.7 & 29.94 & 42.6 & 29.77 & 42.64 & 29.90 \\
& Self-attention mask
& 58.0
& 44.4 & 31.04 & 43.0 & 30.18 & 39.4 & 27.79 & 44.8 & 31.57 & 42.3 & 29.61 & 42.78 & 30.04 \\
& Binary mask (Ours)
& 58.0
& \underline{47.9} & \underline{33.62} & \textbf{46.3} & \textbf{32.63} & \textbf{44.1} & \textbf{30.98} & \textbf{46.9} & \textbf{33.09} & \underline{45.3} & \underline{31.67} & \textbf{46.10} & \textbf{32.40} \\
\multirow{-9}{*}{\enc{VideoMAEv2-B}}
& Soft mask (Ours)
& \na
& \textbf{48.0} & \textbf{33.74} & \underline{46.1} & \underline{32.47} & \underline{44.0} & \underline{30.97} & \underline{46.7} & \underline{32.92} & \textbf{45.4} & \textbf{31.79} & \underline{46.04} & \underline{32.38} \\

\bottomrule
\end{tabular}%
}
\end{table}

\subsection{Downstream Robustness to the Video Generator}
\label{app:generator-robustness}

The previous analysis shows that the difference in generative consistency
between general and specialized concepts is observed across multiple
text-to-video generators. We next investigate whether this property also
results in useful feature masks when the source generator used to construct
the synthetic training set is changed.

To this end, we independently train masks using the synthetic datasets
generated by CogVideoX1.5-5B and Mochi~1 and evaluate them using the same
downstream protocols as the Wan2.1-based masks in the main experiments.
Importantly, only the source of the synthetic mask-training data changes;
the downstream datasets, feature encoders, and evaluation procedures remain
unchanged.

\paragraph{Human similarity alignment.}
Table~\ref{tab:convisbench-gen} evaluates masks learned from
CogVideoX1.5-5B and Mochi~1 on ConViS-Bench. For both V-JEPA 2.1-L and
VideoMAEv2-B, masks learned from either generator improve the average
correlation with human similarity judgements compared with the corresponding
unmasked representations. For V-JEPA 2.1-L, for example, the average
Spearman correlation increases from $31.14$ for the unmasked representation
to $37.20$ and $35.66$ using binary masks derived from soft masks learned on CogVideoX1.5-5B
and Mochi~1, respectively. The corresponding soft masks achieve average
correlations of $38.14$ and $36.78$. A similar trend is observed with
VideoMAEv2-B. These results show that the semantic subspaces recovered by
our objective remain useful when the synthetic training videos are produced by different generators.


\newcommand{\new}[1]{#1}                      

\begin{table}[t]
\centering
\caption{\textbf{Effect of the mask source generator on ConViS-Bench.}
Spearman's $\rho$ and Kendall's $\tau$ correlations with human judgement
($\times100$). \emph{Sp.} denotes the sparsity of the learned mask; soft
masks are dense by construction. Masks learned from synthetic datasets
generated by either CogVideoX1.5-5B or Mochi~1 improve average alignment
with human judgement over the corresponding unmasked representations,
demonstrating robustness to the choice of source video generator.
Best results are \textbf{bolded}, second-best \underline{underlined}.}
\label{tab:convisbench-gen}

\scriptsize
\renewcommand{\arraystretch}{1.25}
\setlength{\tabcolsep}{3.5pt}

\settowidth{\encw}{\textbf{VideoMAEv2-B}}
\addtolength{\encw}{10pt}
\addtolength{\encw}{2\tabcolsep}
\setlength{\encblk}{\ht\strutbox}
\addtolength{\encblk}{\dp\strutbox}
\setlength{\encblk}{6.25\encblk}

\resizebox{\textwidth}{!}{%
\begin{tabular}{
  c
  >{\columncolor{StubMeth}}l
  >{\columncolor{SpBody}}c
  >{\columncolor{GrpABody}}c >{\columncolor{GrpABody}}c
  >{\columncolor{GrpBBody}}c >{\columncolor{GrpBBody}}c
  >{\columncolor{GrpABody}}c >{\columncolor{GrpABody}}c
  >{\columncolor{GrpBBody}}c >{\columncolor{GrpBBody}}c
  >{\columncolor{GrpABody}}c >{\columncolor{GrpABody}}c
  >{\columncolor{AvgBody}}c  >{\columncolor{AvgBody}}c}
\toprule

& & \cellcolor{SpHead}
& \multicolumn{2}{>{\columncolor{GrpAHead}}c}{\textbf{Main Action}}
& \multicolumn{2}{>{\columncolor{GrpBHead}}c}{\textbf{Main Subjects}}
& \multicolumn{2}{>{\columncolor{GrpAHead}}c}{\textbf{Main Objects}}
& \multicolumn{2}{>{\columncolor{GrpBHead}}c}{\textbf{Location}}
& \multicolumn{2}{>{\columncolor{GrpAHead}}c}{\textbf{Actions Order}}
& \multicolumn{2}{>{\columncolor{AvgHead}}c}{\textbf{Average}} \\

\textbf{Encoder}
& \textbf{Method}
& \cellcolor{SpHead}\textbf{Sp.\,(\%)}
& \cellcolor{GrpAMet}$\rho$ & \cellcolor{GrpAMet}$\tau$
& \cellcolor{GrpBMet}$\rho$ & \cellcolor{GrpBMet}$\tau$
& \cellcolor{GrpAMet}$\rho$ & \cellcolor{GrpAMet}$\tau$
& \cellcolor{GrpBMet}$\rho$ & \cellcolor{GrpBMet}$\tau$
& \cellcolor{GrpAMet}$\rho$ & \cellcolor{GrpAMet}$\tau$
& \cellcolor{AvgMet}$\rho$  & \cellcolor{AvgMet}$\tau$ \\

\midrule

& Baseline (no mask)
& \na
& 24.0 & 16.30 & 33.4 & 22.84 & 24.7 & 16.83
& 49.3 & 35.04 & 24.3 & 16.65 & 31.14 & 21.53 \\

& \new{Binary mask (CogVideoX)}
& \new{75.3}
& \new{\underline{30.8}} & \new{\underline{21.02}}
& \new{37.9} & \new{26.12}
& \new{31.4} & \new{21.46}
& \new{51.7} & \new{36.94}
& \new{\underline{34.2}} & \new{\underline{23.67}}
& \new{\underline{37.20}} & \new{\underline{25.84}} \\

& \new{Binary mask (Mochi)}
& \new{72.1}
& \new{28.2} & \new{19.30}
& \new{37.5} & \new{25.91}
& \new{30.1} & \new{20.48}
& \new{52.3} & \new{37.44}
& \new{30.2} & \new{20.77}
& \new{35.66} & \new{24.78} \\

& \new{Soft mask (CogVideoX)}
& \na
& \new{\textbf{31.8}} & \new{\textbf{21.76}}
& \new{\textbf{38.9}} & \new{\textbf{26.94}}
& \new{\textbf{32.3}} & \new{\textbf{22.08}}
& \new{\textbf{53.0}} & \new{\underline{37.88}}
& \new{\textbf{34.7}} & \new{\textbf{24.04}}
& \new{\textbf{38.14}} & \new{\textbf{26.54}} \\

\multirow{-5}{*}{\enc{V-JEPA 2.1-L}}
& \new{Soft mask (Mochi)}
& \na
& \new{29.4} & \new{20.04}
& \new{\underline{38.0}} & \new{\underline{26.26}}
& \new{\underline{31.6}} & \new{\underline{21.47}}
& \new{\underline{53.0}} & \new{\textbf{37.97}}
& \new{31.9} & \new{22.04}
& \new{36.78} & \new{25.56} \\

\midrule

& Baseline (no mask)
& \na
& \underline{47.0} & 33.02
& 44.6 & 31.46
& 42.4 & 29.96
& 45.8 & 32.26
& \textbf{44.7} & \textbf{31.37}
& 44.90 & 31.61 \\

& \new{Binary mask (CogVideoX)}
& \new{57.8}
& \new{46.8} & \new{32.91}
& \new{\textbf{45.9}} & \new{\textbf{32.26}}
& \new{44.0} & \new{30.88}
& \new{\textbf{47.0}} & \new{33.17}
& \new{43.8} & \new{30.59}
& \new{45.50} & \new{31.96} \\

& \new{Binary mask (Mochi)}
& \new{51.9}
& \new{\textbf{47.2}} & \new{\textbf{33.19}}
& \new{45.6} & \new{32.18}
& \new{\textbf{44.5}} & \new{\textbf{31.25}}
& \new{\textbf{47.0}} & \new{\textbf{33.20}}
& \new{\textbf{44.7}} & \new{\underline{31.27}}
& \new{\textbf{45.80}} & \new{\textbf{32.22}} \\

& \new{Soft mask (CogVideoX)}
& \na
& \new{46.9} & \new{32.99}
& \new{\underline{45.7}} & \new{\underline{32.22}}
& \new{43.8} & \new{30.71}
& \new{46.7} & \new{32.94}
& \new{44.0} & \new{30.68}
& \new{45.42} & \new{31.91} \\

\multirow{-5}{*}{\enc{VideoMAEv2-B}}
& \new{Soft mask (Mochi)}
& \na
& \new{46.9} & \new{\underline{33.04}}
& \new{\underline{45.7}} & \new{32.19}
& \new{\underline{44.2}} & \new{\underline{31.08}}
& \new{\underline{46.9}} & \new{\underline{33.19}}
& \new{\underline{44.3}} & \new{31.03}
& \new{\underline{45.60}} & \new{\underline{32.11}} \\

\bottomrule
\end{tabular}%
}
\end{table}

\paragraph{OOD detection.}
We next evaluate whether the generator-specific masks transfer to
video-generation OOD detection. Recall that these masks are learned solely
from synthetic text-to-video generations, whereas the downstream OOD
experiment uses Cosmos-Predict2 for image-to-video generation. This
therefore evaluates transfer not only across datasets but also across
generative tasks and model families.

Table~\ref{tab:OOD-results-cogvideox} reports the results obtained using
masks learned from CogVideoX1.5-5B's generated videos. Across the four OOD datasets, both
feature encoders, and the 2XDM and DDPM-OOD scoring procedures, the learned
masks provide substantial improvements over the unmasked representations
in most evaluation settings. Particularly large gains are observed for
MedVidBench and HAM10000, demonstrating that the generative-consistency
signal extracted from CogVideoX generations transfers effectively to
OOD detection with a different generative model.

\begin{table}[t]
\centering
\caption{\textbf{OOD detection using masks learned from CogVideoX1.5-5B.}
Video generation OOD detection across four datasets using Cosmos-Predict2,
with BDD100k as the in-distribution (ID) baseline. The soft masks are learned exclusively from synthetic videos generated by
CogVideoX1.5-5B, while the corresponding binary masks are obtained by post-hoc thresholding at $\theta=0.5$.
The resulting improvements over the unmasked representations demonstrate
transfer across source video generators and generative tasks.
Best results are \textbf{bolded} and second-best results are
\underline{underlined}.}
\label{tab:OOD-results-cogvideox}

\scriptsize
\renewcommand{\arraystretch}{1.2}
\setlength{\tabcolsep}{3.5pt}

\resizebox{\textwidth}{!}{%
\begin{tabular}{l l
  >{\columncolor{XDMBody}}c >{\columncolor{XDMBody}}c >{\columncolor{XDMBody}}c
  >{\columncolor{DDPMBody}}c >{\columncolor{DDPMBody}}c >{\columncolor{DDPMBody}[\tabcolsep][0pt]}c
  >{\columncolor{XDMBody}[0pt][\tabcolsep]}c >{\columncolor{XDMBody}}c >{\columncolor{XDMBody}}c
  >{\columncolor{DDPMBody}}c >{\columncolor{DDPMBody}}c >{\columncolor{DDPMBody}}c}
\toprule

& &
\multicolumn{6}{>{\columncolor{EncoderBlue}[\tabcolsep][0pt]}c}
{\textcolor{white}{\textbf{V-JEPA 2.1-L}}} &
\multicolumn{6}{>{\columncolor{EncoderBlue}[0pt][\tabcolsep]}c}
{\textcolor{white}{\textbf{VideoMAEv2-B}}} \\[1pt]

& &
\multicolumn{3}{>{\columncolor{XDMHead}}c}{\textbf{2XDM}} &
\multicolumn{3}{>{\columncolor{DDPMHead}[\tabcolsep][0pt]}c}{\textbf{DDPM-OOD}} &
\multicolumn{3}{>{\columncolor{XDMHead}[0pt][\tabcolsep]}c}{\textbf{2XDM}} &
\multicolumn{3}{>{\columncolor{DDPMHead}}c}{\textbf{DDPM-OOD}} \\

\multirow{-3}{*}{\textbf{OOD Dataset}} &
\multirow{-3}{*}{\textbf{Method}} &
\cellcolor{XDMMetric}AUROC\,$\uparrow$ &
\cellcolor{XDMMetric}AUPR\,$\uparrow$ &
\cellcolor{XDMMetric}FPR@95\,$\downarrow$ &
\cellcolor{DDPMMetric}AUROC\,$\uparrow$ &
\cellcolor{DDPMMetric}AUPR\,$\uparrow$ &
\multicolumn{1}{>{\columncolor{DDPMMetric}[\tabcolsep][0pt]}c}
{FPR@95\,$\downarrow$} &
\multicolumn{1}{>{\columncolor{XDMMetric}[0pt][\tabcolsep]}c}
{AUROC\,$\uparrow$} &
\cellcolor{XDMMetric}AUPR\,$\uparrow$ &
\cellcolor{XDMMetric}FPR@95\,$\downarrow$ &
\cellcolor{DDPMMetric}AUROC\,$\uparrow$ &
\cellcolor{DDPMMetric}AUPR\,$\uparrow$ &
\cellcolor{DDPMMetric}FPR@95\,$\downarrow$ \\

\midrule

\multirow{3}{*}{\textbf{UVEB}}
& Unmasked
& 78.42 & 73.77 & \underline{57.20}
& \underline{76.37} & 70.31 & 63.60
& 71.08 & 68.50 & \underline{77.60}
& 75.50 & 73.33 & \textbf{83.20} \\

& Binary mask
& \underline{84.19} & \underline{82.29} & \textbf{52.80}
& \textbf{89.43} & \underline{88.51} & \underline{42.40}
& \textbf{73.18} & \textbf{70.24} & \textbf{71.20}
& \textbf{78.97} & \textbf{76.73} & \underline{84.40} \\

& Soft mask
& \textbf{84.63} & \textbf{82.98} & \textbf{52.80}
& \textbf{89.43} & \textbf{88.54} & \textbf{42.00}
& \underline{72.98} & \underline{70.03} & \textbf{71.20}
& \underline{78.55} & \underline{76.38} & \underline{84.40} \\

\midrule

\multirow{3}{*}{\textbf{MedVidBench}}
& Unmasked
& 74.66 & 74.18 & 76.00
& 68.11 & 68.47 & \underline{86.00}
& 78.59 & 80.73 & 82.40
& 77.01 & 79.46 & 79.60 \\

& Binary mask
& \underline{89.47} & \underline{90.36} & \textbf{58.80}
& \underline{92.99} & \underline{93.23} & \textbf{34.00}
& \textbf{83.00} & \textbf{84.40} & \textbf{73.60}
& \textbf{81.85} & \underline{83.53} & \underline{78.40} \\

& Soft mask
& \textbf{89.81} & \textbf{90.64} & \underline{59.20}
& \textbf{93.10} & \textbf{93.31} & \textbf{34.00}
& \underline{82.87} & \underline{84.30} & \underline{74.40}
& \underline{81.72} & \textbf{83.59} & \textbf{77.20} \\

\midrule

\multirow{3}{*}{\textbf{HAM10000}}
& Unmasked
& 70.80 & 61.21 & \underline{63.20}
& \underline{87.67} & \underline{87.95} & 51.60
& 68.38 & 62.42 & \underline{78.40}
& 86.23 & 88.37 & \textbf{67.60} \\

& Binary mask
& \underline{79.52} & \underline{72.17} & \textbf{55.20}
& \textbf{96.43} & \textbf{96.90} & \underline{19.60}
& \textbf{70.41} & \textbf{63.77} & \textbf{74.40}
& \textbf{88.66} & \textbf{90.32} & \underline{68.80} \\

& Soft mask
& \textbf{80.01} & \textbf{73.22} & \textbf{55.20}
& \textbf{96.43} & \textbf{96.90} & \textbf{19.20}
& \underline{70.27} & \underline{63.74} & \textbf{74.40}
& \underline{88.51} & \underline{90.22} & 69.20 \\

\midrule

\multirow{3}{*}{\textbf{WikiArt}}
& Unmasked
& 63.10 & 55.82 & \textbf{65.20}
& 69.80 & 65.03 & 82.40
& 57.78 & 51.05 & \underline{82.00}
& 63.88 & 63.16 & \underline{89.20} \\

& Binary mask
& \underline{66.22} & \underline{59.83} & \underline{66.80}
& \underline{80.75} & \underline{76.70} & \underline{68.80}
& \textbf{59.11} & \textbf{51.66} & \textbf{78.40}
& \textbf{66.78} & \textbf{64.97} & \textbf{88.00} \\

& Soft mask
& \textbf{67.21} & \textbf{61.13} & \underline{66.80}
& \textbf{81.00} & \textbf{76.73} & \textbf{67.60}
& \underline{58.91} & \underline{51.54} & \textbf{78.40}
& \underline{66.32} & \underline{64.52} & \textbf{88.00} \\

\bottomrule
\end{tabular}%
}
\end{table}

Table~\ref{tab:OOD-results-mochi} repeats the same evaluation using masks
learned from Mochi~1. The overall behavior is consistent with the
CogVideoX1.5-5B results: masking improves OOD discrimination across a broad
range of datasets, feature encoders, and scoring procedures, with especially
clear improvements on MedVidBench and HAM10000. Although the magnitude of
the gains varies with the source generator and downstream configuration,
the masks learned from both alternative generators remain substantially
more effective than the corresponding unmasked representations in the
majority of settings.

\paragraph{Retrieval.} We extend video retrieval experiments on UCF101 and HMDB51 with the two new video generators.
Results in Table~\ref{tab:retrieval_mochi_cogvideox} are consistent with those reported in Table~\ref{tab:retrieval_wan} for Wan2.1, with actually stronger performances given by masks learned on CogVideoX videos (+3.6 and +2.3 mAP for V-JEPA features).

\begin{table*}[t]
  \centering
  \caption{Class-level \textbf{video retrieval mAP (\%)} on UCF101 and HMDB51,
    without and with the masks learned from Mochi 1 and CogVideoX generated datasets.
    \textit{Sp.} denotes mask sparsity in \%.
    Best results are \textbf{bolded} and second-best results are
\underline{underlined} per generator.}
  \label{tab:retrieval_mochi_cogvideox}
  \setlength{\tabcolsep}{8pt}
  \renewcommand{\arraystretch}{1.22}
  \resizebox{\textwidth}{!}{%
  \begin{tabular}{l l c ccc ccc}
    \arrayrulecolor{hdr}\specialrule{1.1pt}{0pt}{0pt}
    \rowcolor{hdr}
    \cellcolor{hdr}\gp & \cellcolor{hdr} & \cellcolor{hdr} &
    \multicolumn{3}{c}{\cellcolor{hdr}\color{white}\bfseries Mochi 1} &
    \multicolumn{3}{c}{\cellcolor{hdr}\color{white}\bfseries CogVideoX1.5-5B} \\
    \arrayrulecolor{white}\cmidrule(lr){4-6}\cmidrule(lr){7-9}\arrayrulecolor{hdr}
    \rowcolor{hdr}
    \color{white}\bfseries Dataset & \cellcolor{hdr}\color{white}\bfseries Encoder &
    \cellcolor{hdr}\color{white}\bfseries Unmasked &
    \color{white}Sp. (\%) & \color{white}Binary & \color{white}Soft &
    \color{white}Sp. (\%) & \color{white}Binary & \color{white}Soft \\
    \specialrule{1.1pt}{0pt}{0pt}\arrayrulecolor{black}

    \multirow{2}{*}{\dset{UCF101}}
      & V-JEPA 2.1 Large & 40.82 & \sparsity{72.2} & \runr{42.84} & \best{43.48}
                                 & \sparsity{75.4} & \runr{44.02} & \best{44.42} \\
      & VideoMAEv2-Base  & 95.52 & \sparsity{52.0} & \runr{95.56} & \best{95.62}
                                 & \sparsity{57.8} & \runr{95.55} & \best{95.61} \\
    \arrayrulecolor{black!25}\midrule\arrayrulecolor{black}
    \multirow{2}{*}{\dset{HMDB51}}
      & V-JEPA 2.1 Large & 17.84 & \sparsity{72.2} & \runr{18.99} & \best{19.59}
                                 & \sparsity{75.4} & \runr{19.91} & \best{20.19} \\
      & VideoMAEv2-Base  & \best{46.94} & \sparsity{52.0} & 46.59 & \runr{46.88}
                                 & \sparsity{57.8} & 46.51 & \runr{46.74} \\
    \arrayrulecolor{hdr}\specialrule{1.1pt}{2pt}{0pt}\arrayrulecolor{black}
  \end{tabular}}
\end{table*}

\paragraph{Overall robustness results.}
Together with the Wan2.1 results in the main paper, these experiments show
that the effectiveness of our learned feature subspaces does not depend on
using a particular T2V model to construct the synthetic training dataset.
Wan2.1, CogVideoX1.5-5B, and Mochi~1 differ in architecture, training data and objective,
yet the same general-versus-specialized consistency signal produces masks
that transfer to downstream semantic-similarity and OOD-detection tasks.

\begin{table}[t]
\centering
\caption{\textbf{OOD detection using masks learned from Mochi 1.}
Video generation OOD detection across four datasets using Cosmos-Predict2,
with BDD100k as the in-distribution (ID) baseline. The soft masks are learned exclusively from synthetic videos generated by
Mochi 1, while the corresponding binary masks are obtained by post-hoc thresholding at $\theta=0.5$. The
improvements across diverse OOD domains provide additional evidence that
the proposed learning signal transfers across source video generators.
Best results are \textbf{bolded} and second-best results are
\underline{underlined}.}
\label{tab:OOD-results-mochi}

\scriptsize
\renewcommand{\arraystretch}{1.2}
\setlength{\tabcolsep}{3.5pt}

\resizebox{\textwidth}{!}{%
\begin{tabular}{l l
  >{\columncolor{XDMBody}}c >{\columncolor{XDMBody}}c >{\columncolor{XDMBody}}c
  >{\columncolor{DDPMBody}}c >{\columncolor{DDPMBody}}c >{\columncolor{DDPMBody}[\tabcolsep][0pt]}c
  >{\columncolor{XDMBody}[0pt][\tabcolsep]}c >{\columncolor{XDMBody}}c >{\columncolor{XDMBody}}c
  >{\columncolor{DDPMBody}}c >{\columncolor{DDPMBody}}c >{\columncolor{DDPMBody}}c}
\toprule

& &
\multicolumn{6}{>{\columncolor{EncoderBlue}[\tabcolsep][0pt]}c}
{\textcolor{white}{\textbf{V-JEPA2.1-L}}} &
\multicolumn{6}{>{\columncolor{EncoderBlue}[0pt][\tabcolsep]}c}
{\textcolor{white}{\textbf{VideoMAEv2-B}}} \\[1pt]

& &
\multicolumn{3}{>{\columncolor{XDMHead}}c}{\textbf{2XDM}} &
\multicolumn{3}{>{\columncolor{DDPMHead}[\tabcolsep][0pt]}c}{\textbf{DDPM-OOD}} &
\multicolumn{3}{>{\columncolor{XDMHead}[0pt][\tabcolsep]}c}{\textbf{2XDM}} &
\multicolumn{3}{>{\columncolor{DDPMHead}}c}{\textbf{DDPM-OOD}} \\

\multirow{-3}{*}{\textbf{OOD Dataset}} &
\multirow{-3}{*}{\textbf{Method}} &
\cellcolor{XDMMetric}AUROC\,$\uparrow$ &
\cellcolor{XDMMetric}AUPR\,$\uparrow$ &
\cellcolor{XDMMetric}FPR@95\,$\downarrow$ &
\cellcolor{DDPMMetric}AUROC\,$\uparrow$ &
\cellcolor{DDPMMetric}AUPR\,$\uparrow$ &
\multicolumn{1}{>{\columncolor{DDPMMetric}[\tabcolsep][0pt]}c}
{FPR@95\,$\downarrow$} &
\multicolumn{1}{>{\columncolor{XDMMetric}[0pt][\tabcolsep]}c}
{AUROC\,$\uparrow$} &
\cellcolor{XDMMetric}AUPR\,$\uparrow$ &
\cellcolor{XDMMetric}FPR@95\,$\downarrow$ &
\cellcolor{DDPMMetric}AUROC\,$\uparrow$ &
\cellcolor{DDPMMetric}AUPR\,$\uparrow$ &
\cellcolor{DDPMMetric}FPR@95\,$\downarrow$ \\

\midrule

\multirow{3}{*}{\textbf{UVEB}}
& Unmasked
& 78.42 & 73.77 & 57.20
& 76.37 & 70.31 & 63.60
& 71.08 & 68.50 & 77.60
& 75.50 & 73.33 & \textbf{83.20} \\

& Binary mask
& \underline{81.84} & \underline{78.98} & \underline{54.00}
& \underline{84.85} & \underline{81.37} & \underline{51.20}
& \textbf{72.32} & \textbf{69.96} & \underline{75.20}
& \textbf{78.01} & \textbf{76.01} & \textbf{83.20} \\

& Soft mask
& \textbf{82.61} & \textbf{80.33} & \textbf{52.80}
& \textbf{85.99} & \textbf{83.20} & \textbf{48.80}
& \underline{72.19} & \underline{69.92} & \textbf{74.80}
& \underline{77.66} & \underline{75.76} & \underline{83.60} \\

\midrule

\multirow{3}{*}{\textbf{MedVidBench}}
& Unmasked
& 74.66 & 74.18 & 76.00
& 68.11 & 68.47 & 86.00
& 78.59 & 80.73 & 82.40
& 77.01 & 79.46 & \textbf{79.60} \\

& Binary mask
& \underline{85.93} & \underline{86.71} & \underline{64.40}
& \underline{88.74} & \underline{88.54} & \underline{55.60}
& \textbf{80.69} & \textbf{82.01} & \underline{82.00}
& \textbf{79.97} & \textbf{82.25} & 80.80 \\

& Soft mask
& \textbf{86.74} & \textbf{87.56} & \textbf{62.80}
& \textbf{90.12} & \textbf{90.26} & \textbf{52.00}
& \underline{80.45} & \underline{81.85} & \textbf{80.40}
& \underline{79.54} & \underline{81.97} & \underline{80.00} \\

\midrule

\multirow{3}{*}{\textbf{HAM10000}}
& Unmasked
& 70.80 & 61.21 & 63.20
& 87.67 & 87.95 & 51.60
& 68.38 & 62.42 & 78.40
& 86.23 & 88.37 & 67.60 \\

& Binary mask
& \underline{76.42} & \underline{68.06} & \underline{60.40}
& \underline{93.72} & \underline{94.16} & \underline{32.00}
& \textbf{71.00} & \textbf{64.98} & \underline{76.40}
& \textbf{89.46} & \textbf{91.01} & \textbf{60.40} \\

& Soft mask
& \textbf{77.18} & \textbf{69.06} & \textbf{58.40}
& \textbf{94.32} & \textbf{94.79} & \textbf{30.00}
& \underline{70.79} & \underline{64.80} & \textbf{76.00}
& \underline{89.00} & \underline{90.63} & \underline{62.00} \\

\midrule

\multirow{3}{*}{\textbf{WikiArt}}
& Unmasked
& 63.10 & 55.82 & 65.20
& 69.80 & 65.03 & 82.40
& 57.78 & 51.05 & \textbf{82.00}
& 63.88 & 63.16 & 89.20 \\

& Binary mask
& \underline{66.87} & \underline{60.39} & \textbf{64.40}
& \underline{76.14} & \underline{70.05} & \underline{72.00}
& \textbf{58.45} & \textbf{51.34} & \underline{82.40}
& \textbf{65.88} & \textbf{64.35} & \textbf{86.80} \\

& Soft mask
& \textbf{67.20} & \textbf{60.77} & \underline{64.80}
& \textbf{77.39} & \textbf{71.96} & \textbf{69.20}
& \underline{57.99} & \underline{51.08} & 82.80
& \underline{64.95} & \underline{63.46} & \underline{88.40} \\

\bottomrule
\end{tabular}%
}
\end{table}

\subsection{Extending global masks to dense local tasks}\label{sec:vos-tracking}

While our mask is optimized using global representations, we evaluate its transferability to dense, patch-level tasks where similarity is computed at a finer granularity. Following~\cite{mur2026v}, we assess performance on Video Object Segmentation (VOS) using the DAVIS-2017 dataset~\citep{Davis2017}. In this task, an initial ground-truth object mask must be propagated across subsequent frames. 
We adopt the non-parametric label propagation approach detailed in~\cite{mur2026v}, which matches local patch features between frames via cosine similarity in the V-JEPA 2.1 embedding space. To ensure a fair comparison, we maintain the hyperparameters reported in the original study.

\begin{figure*}[h]
  \centering
    \includegraphics[width=0.7\linewidth]{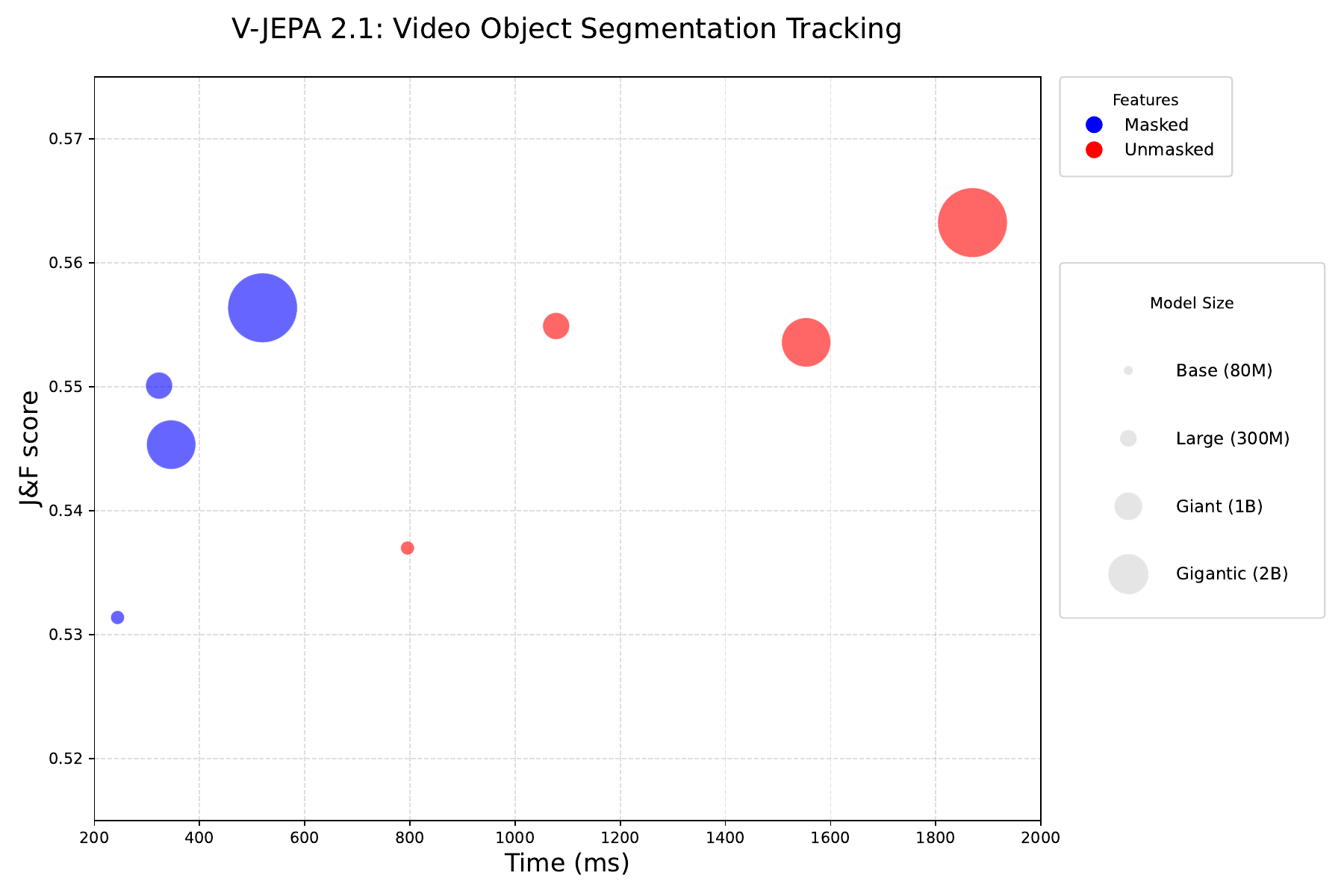} 
    \caption{\textbf{Patch-level effectiveness of masked features.} VOS on DAVIS-2017~\citep{Davis2017} via patch features matching.
    }
    \label{fig:imgs}
\end{figure*}

\textbf{Efficient dense tracking with masked features.} We evaluate all model variants using the full feature set as a baseline, and then repeat the experiments by applying our learned mask to prune the feature dimensions. As illustrated in Fig.~\ref{fig:imgs}, our masked subspace maintains competitive segmentation performance while providing a significant computational advantage. Specifically, by reducing the dimensionality of the search space, our approach accelerates the label propagation algorithm by $\approx$\textbf{4$\times$}, demonstrating that the semantic features identified at the global level remain highly effective for localized tracking.

For our experiments, we use V-JEPA 2.1 with all model sizes and follow the best hyperparameters reported by~\cite{mur2026v}, i.e., a context window of 15 frames, a spatial search radius of 12, $k=5$ nearest neighbors, and a softmax temperature of $0.2$. Reported time refers to the inference performed on a single NVIDIA H100 GPU.

\subsection{Full details on reconstruction via patch clustering}\label{app:reconstruction}
Following prior work on analyzing the spatial structure encoded in visual representations \citep{DHANACHANDRA2015764}, we assess whether our masked features preserve local visual information. For each video clip in the dataset, we extract a single RGB frame $I \in \mathbb{R}^{3 \times H \times W}$ and compute patch-level feature tokens $z_i \in \mathbb{R}^d$ using the frozen V-JEPA 2.1 encoder with our learned mask applied. We then perform K-means clustering on the set of patch tokens $\{z_i\}$ in the embedding space, partitioning them into $k$ clusters. By assigning each patch token's cluster label back to its original spatial location, we obtain a piecewise-constant segmentation of the image in which each pixel $p$ is associated with a discrete cluster label $\ell(p) \in \{1, \dots, k\}$.

We then generate a mosaic reconstruction $\hat{I}$ of the original frame by replacing the RGB value of each pixel with the mean color vector of all pixels sharing the same cluster assignment. The reconstruction error for a single frame is quantified by computing the Mean Squared Error (MSE) between the original frame $I$ and the reconstruction $\hat{I}$, averaged over all spatial locations and color channels:
\begin{equation}
    \operatorname{MSE}_k(I) = \frac{1}{3HW} \sum_{c,h,w} (I_{c,h,w} - \hat{I}_{c,h,w})^2
\end{equation}

\begin{figure}[h!]
    \centering
    \includegraphics[width=\linewidth]{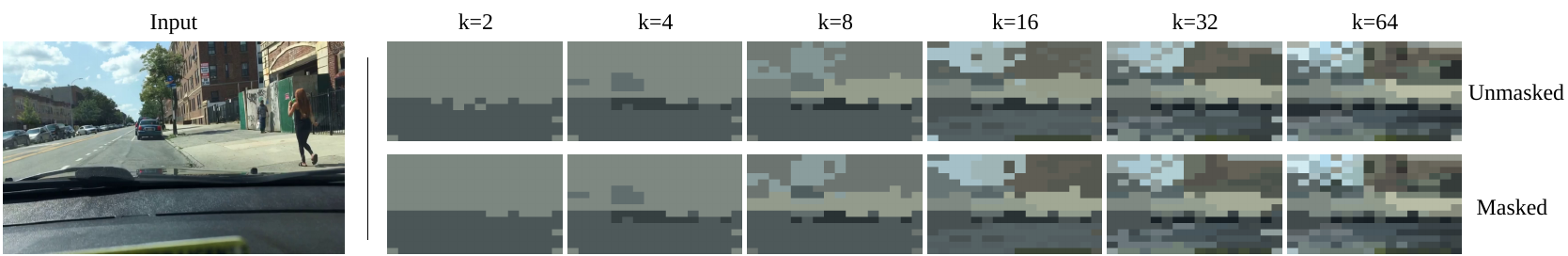}
    \caption{\textbf{Qualitative visualization of K-means mosaic reconstructions.} The original input frame (left) is segmented based on patch-level cluster assignments for $k \in \{2, 4, 8, 16, 32, 64\}$. The visual consistency between the unmasked baseline (top row) and our masked representations (bottom row) illustrates that the masked feature space preserves essential local structural and color information.}
    \label{fig:kmeans_mosaic_threshold_viz}
\end{figure}

To qualitatively assess this preservation, Figure \ref{fig:kmeans_mosaic_threshold_viz} visualizes the mosaic reconstructions across varying cluster counts ($k$), demonstrating that the spatial and color regions produced by the masked representations closely mirror the unmasked baseline. Quantitatively, Figure \ref{fig:kmeans_mosaic_threshold} reports the final $\operatorname{MSE}_k$ averaged across all evaluated frames. We compare the unmasked features against our learned mask at varying binarization thresholds ($\theta \in \{0.5, 0.7, 0.9\}$). The average reconstruction MSE for the masked features remains remarkably close to the full baseline across all evaluated values of $k$ and $\theta$, confirming that the heavily pruned subspaces retain the core visual features.

\subsection{Broader impacts} \label{sec:impacts}
This research work proposes a lightweight model-agnostic method to learn a semantic similarity score on top of pretrained feature encoders. Such similarity shows improvements in alignment with human judgment and, as shown with experiments over various tasks, can be used for a variety of real-world use cases that rely on the concept of similarity between two videos. 
Additionally, it achieves such results with higher efficiency than baseline methods by effectively reducing the representation dimensions, and thus the storage required by the extracted features.
Negative impacts could result from the limitations and biases inherited from the models used for data synthesis and feature encoding.

\section{Limitations} While our framework demonstrates the utility of generative uncertainty for representation adaptation, several limitations remain. 
First, our pipeline is inherently coupled with the performance and inductive biases of the underlying pre-trained components, including Gemini for prompt synthesis, video generative models for video generation, and the foundation models used as encoders. Any systematic biases or domain-specific shortcomings in these models, such as cultural biases in text-to-video synthesis, may propagate into the learned similarity masks. 
Second, our optimization objective prioritizes global semantic consistency. While this yields highly compact and efficient representations, we observed a marginal performance trade-off in tasks requiring dense, fine-grained local precision (e.g., video object segmentation). Future work could explore multi-scale masking strategies to better bridge this gap. 

\section{Licenses}\label{sec:licenses}
List of licenses for existing assets used in this paper.

\textbf{Models:}
\begin{itemize}
    \item Wan2.1 (T2V-1.3B): Apache License Version 2.0
    \item CogVideoX1.5-5B: CogVideoX License
    \item Mochi 1: Apache License Version 2.0
    \item V-JEPA 2.1: MIT license
    \item VideoMAEv2: CC-BY-NC-4.0 license
    \item Cosmos-predict2: Apache License Version 2.0
\end{itemize}

\textbf{Data:}
\begin{itemize}
    \item VBench: Apache License Version 2.0
    \item BDD100k: BSD 3-Clause License
    \item UVEB: MIT License
    \item MedVidBench: CC-BY-NC-SA-4.0
    \item HAM10000: CC-BY-NC-SA-4.0
    \item WikiArt: CC0: Public Domain
    \item DAVIS-2017: BSD 3-Clause License
    \item UCF-101: CC0: Public Domain
    \item HMDB51: CC BY 4.0
\end{itemize}